\pdfoutput=1

\documentclass[11pt]{article}

\usepackage{ACL2023}

\usepackage{times}
\usepackage{latexsym}

\usepackage[T1]{fontenc}

\usepackage[utf8]{inputenc}

\usepackage{microtype}

\usepackage{inconsolata}

\usepackage{booktabs}
\usepackage{xspace}
\usepackage{graphicx}
\usepackage{float}
\usepackage{url}
\usepackage{listings}
\usepackage{adjustbox}
\usepackage{xcolor}
\usepackage{algpseudocode}
\usepackage{longtable}
\usepackage[most]{tcolorbox}
\usepackage{mdframed}
\usepackage{lipsum}
\usepackage{multirow}
\usepackage{amsfonts}
\usepackage{rotating}
\usepackage{enumitem}

\newcommand*{\img}[1]{%
    \raisebox{-.40\baselineskip}{%
        \includegraphics[
        width=3.2\baselineskip,
        keepaspectratio,
        ]{#1}%
    }%
}

\newcommand{\icon}{\img{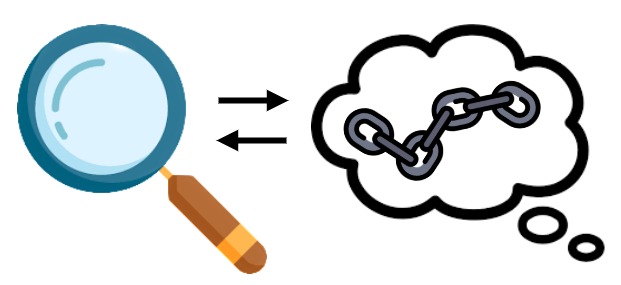}\xspace} 
\newcommand{\fixedicon}{} 
\newcommand{\sys}{\underline{IRCoT}\xspace} 

\newcommand{\iconsys}{\sys} 
\newcommand{\p}[1]{\phantom{#1}}
\newcommand{\nf}[1]{#1}
\newcommand{\std}[1]{\scriptsize{$\pm$ #1}}

\title{\icon Interleaving Retrieval with Chain-of-Thought Reasoning\\ for Knowledge-Intensive Multi-Step Questions}

\newcommand{\affa}{$^\dagger$}
\newcommand{\affb}{$^\ddagger$}

\author{
  Harsh Trivedi\affa \ \ \ Niranjan Balasubramanian\affa \ \ \\
  \\
  \affa Stony Brook University\\
  Stony Brook, U.S.A.\\
  \texttt{\small \{hjtrivedi,niranjan\}@cs.stonybrook.edu}
  \And
  \hspace{5ex} Tushar Khot\affb \ \ \ Ashish Sabharwal\affb \\
  \\
  \hspace{5ex} \affb Allen Institute for AI\\
  \hspace{5ex} Seattle, U.S.A.\\
  \hspace{5ex} \texttt{\small \{tushark,ashishs\}@allenai.org}
}

\begin{document}

\maketitle

\begin{abstract}
Prompting-based large language models (LLMs) are surprisingly powerful at generating natural language reasoning steps or Chains-of-Thoughts (CoT) for multi-step question answering (QA). They struggle, however, when the necessary knowledge is either unavailable to the LLM or not up-to-date within its parameters. While using the question to retrieve relevant text from an external knowledge source helps LLMs, we observe that this one-step retrieve-and-read approach is insufficient for multi-step QA. Here, \textit{what to retrieve} depends on \textit{what has already been derived}, which in turn may depend on \textit{what was previously retrieved}. To address this, we propose IRCoT, a new approach for multi-step QA that interleaves retrieval with steps (sentences) in a CoT, guiding the retrieval with CoT and in turn using retrieved results to improve CoT. Using IRCoT with GPT3 substantially improves retrieval (up to 21 points) as well as downstream QA (up to 15 points) on four datasets: HotpotQA, 2WikiMultihopQA, MuSiQue, and IIRC. We observe similar substantial gains in out-of-distribution (OOD) settings as well as with much smaller models such as Flan-T5-large without additional training. IRCoT reduces model hallucination, resulting in factually more accurate CoT reasoning.\footnote{Code, data, and prompts are available at \url{https://github.com/stonybrooknlp/ircot}}.
\end{abstract}

\section{Introduction}

\begin{figure}[t]
\centering
\includegraphics[width=0.45\textwidth]{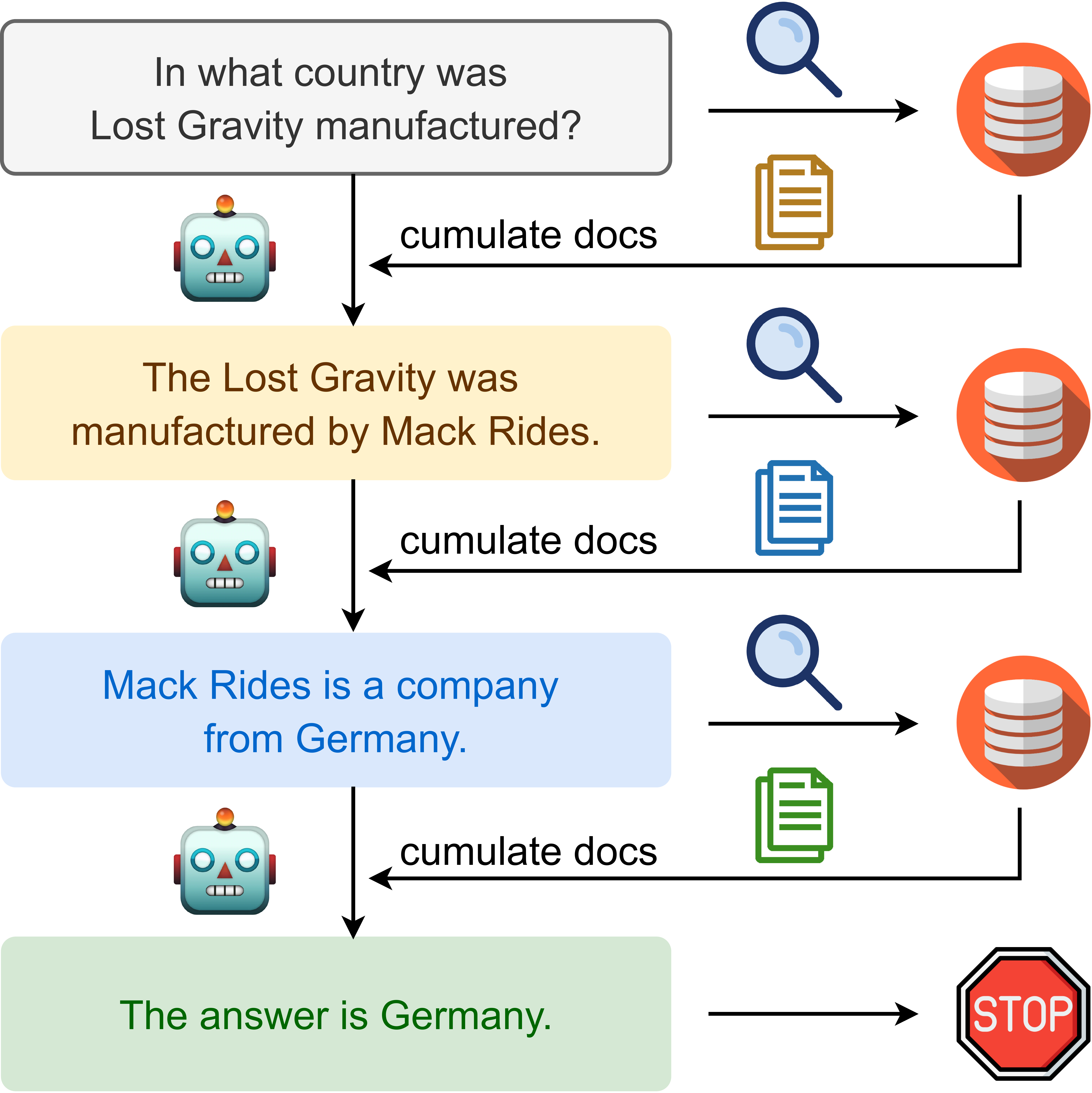}
\caption{
\iconsys interleaves chain-of-thought (CoT) generation and knowledge retrieval steps in order to guide the retrieval by CoT and vice-versa. This interleaving allows retrieving more relevant information for later reasoning steps, compared to standard retrieval using solely the question as the query.}
\label{fig:intro-figure}
\end{figure}

Large language models are capable of answering complex questions by generating step-by-step natural language reasoning steps---so called chains of thoughts (CoT)---when prompted appropriately~\cite{cot}. This approach has been successful when all information needed to answer the question is either provided as context (e.g., algebra questions) or assumed to be present in the model's parameters (e.g., commonsense reasoning). However, for many open-domain questions, all required knowledge is not always available or up-to-date in models' parameters and it's beneficial to retrieve knowledge from external sources~\cite{internet-augmented-qa,realtime-qa}.

\emph{How can we augment chain-of-thought prompting for open-domain, knowledge-intensive tasks that require complex, multi-step reasoning?}

While a \textit{one-shot} retrieval from a knowledge source based solely on the question can successfully augment LMs with relevant knowledge for many factoid-based tasks~\cite{rag,realm,retro,atlas}, this strategy has clear limitations for more complex multi-step reasoning questions. For such questions, one often must retrieve partial knowledge, perform partial reasoning, retrieve additional information based on the outcome of the partial reasoning done so far, and iterate. 
As an example, consider the question illustrated in Fig.~\ref{fig:intro-figure}, \textit{``In what country was Lost Gravity manufactured?''}. The Wikipedia document retrieved using the question (in particular, the roller coaster Lost Gravity) as the query does not mention where Lost Gravity was manufactured. Instead, one must first infer that it was manufactured by a company called Mack Rides, and then perform further retrieval, guided by the inferred company name, to obtain evidence pointing to the manufacturing country.

Thus, the retrieval and reasoning steps must inform each other. Without retrieval, a model is likely to generate an incorrect reasoning step due to hallucination. Additionally, without generating the first reasoning step, the text supporting the second step can't be identified easily given the lack of lexical or even semantic overlap with the question. In other words, we need retrieved facts in order to generate factually correct reasoning steps and the reasoning steps to retrieve relevant facts.

Based on this intuition, we propose an \emph{interleaving approach} to this problem, where the idea is to use retrieval to guide the chain-of-thought (CoT) reasoning steps and use CoT reasoning to guide the retrieval. Fig.~\ref{fig:intro-figure} shows an overview of our retrieval method, which we call \iconsys.\footnote{\underline{I}nterleaved \underline{R}etrieval guided by \underline{C}hain-\underline{o}f-\underline{T}hought.} We begin by retrieving a base set of paragraphs using the question as a query. Subsequently, we alternate between the following two steps: (i) \textit{extend CoT}: use the question, the paragraphs collected thus far, and the CoT sentences generated thus far to generate the next CoT sentence; (ii) \textit{expand retrieved information}: use the last CoT sentence as a query to retrieve additional paragraphs to add to the collected set. We repeat these steps till the CoT reports an answer or we reach the maximum allowed number of reasoning steps. Upon termination, all collected paragraphs are returned as the retrieval outcome. Finally, we use these as the context for answering the question via direct QA prompting~\cite{originalgpt3} or CoT prompting~\cite{cot}.

We evaluate the efficacy of our system on 4 multi-step reasoning datasets under an open-domain setting: HotpotQA~\cite{hotpotqa}, 2WikiMultihopQA~\cite{xanh2020_2wikimultihop}, MuSiQue~\cite{musique}, and IIRC~\cite{iirc}. Our experiments using OpenAI GPT3 (\texttt{code-davinci-002}) \cite{originalgpt3,instructgpt3,codex} demonstrate that retrieval using \iconsys is substantially more effective than the baseline, one-step, question-based retrieval by 11-21 recall points under a fixed-budget optimal recall setup.\footnote{We explain later (in the Metric section and Footnote~\ref{footnote:retrieval-metric}) the appropriateness of this metric in our setting as opposed to more mainstream information recall metrics.} When \iconsys is used in conjunction with a prompting-based reader, it also leads to substantial improvement (up to 15 F1 points) in downstream few-shot QA performance and reduces factual errors in generated CoT by up to 50\%. Our approach also works on much smaller Flan-T5 models (11B, 3B, and 0.7B) showing similar trends. In particular, we find QA using Flan-T5-XL (3B) with \iconsys even outperforms the 58X larger GPT3 with a one-step question-based retrieval.
Furthermore, these improvements also hold up in an out-of-distribution (OOD) setting where the demonstrations from one dataset are used when testing on another dataset.
Lastly, we note that our QA scores exceed those reported by recent works on few-shot prompting for open-domain QA (ODQA)~\cite{decomp,selfask,react}, although a fair apples-to-apples comparison with them isn't possible (cf.~Appendix~\ref{sec:sota-differences}).

In summary, our main \textbf{contribution} is a novel retrieval method, \iconsys, that leverages LMs' chain-of-thought generation capabilities to guide retrieval and uses retrieval in turn to improve CoT reasoning. We demonstrate that \iconsys:
\begin{enumerate}[nosep]

    \item improves both retrieval and few-shot QA performance on several multi-step open-domain QA datasets, in both IID and OOD settings;

    \item reduces factual errors in generated CoTs; and

    \item improves performance with both large-scale (175B models) as well as smaller-scale models (Flan-T5-*, $\le$11B) without any training.

\end{enumerate}

\section{Related Work}
\label{sec:related_work}

\paragraph{Prompting for Open-Domain QA.}

LLMs can learn various tasks by simply using a few examples as prompts~\cite{originalgpt3}. They've also been shown to answer complex questions by producing step-by-step reasoning (chain-of-thoughts, or CoT) when prompted with a few or zero demonstrations~\cite{cot, zerocot}. Prompting has been applied to open-domain QA~\cite{internet-augmented-qa,recitationlm,Yu2022GenerateRT} but its value in improving retrieval and QA for multi-step open-domain questions remains relatively underexplored.

Recently three approaches have been proposed for multi-step open-domain QA. SelfAsk~\cite{selfask} prompts LLMs to decompose a question into subquestions and answers subquestions by a call to Google Search API. DecomP~\cite{decomp} is a general framework that decomposes a task and delegates sub-tasks to appropriate sub-models. They also decompose questions but delegate retrieval to a BM25-based retriever. Both of these approaches are not developed for CoT reasoning, do not focus on the retrieval problem, and require a single-hop QA model to answer the decomposed questions. Recently proposed ReAct~\cite{react} system frames the problem as generating a sequence of reasoning and action steps. These steps are much more complex, rely on much larger models (PaLM-540B), and require fine-tuning to outperform CoT for multi-step ODQA. Furthermore, none of these works have been shown to be effective for smaller models without any training. While a direct comparison with these approaches is not straightforward (difference in knowledge corpus, LLMs, examples), we find that our ODQA performance is much higher than all their reported numbers where available (\S\ref{sec:exp-results}).

\paragraph{Supervised Multi-Step Open-Domain QA.}
Prior work has explored iterative retrieval for open-domain QA in a fully supervised setting. \citet{multi-step-retriever-reader} proposes an iterative retrieval model that retrieves using a neural query representation and then updates it based on a reading comprehension model's output. \citet{multihop-retriever-odqa} apply similar neural query reformulation idea for multihop open-domain QA. \citet{mpr} extends the widely-used Dense Passage Retrieval (DPR)~\cite{dpr} to multihop setting, which has since been improved by \citet{baleen}. \citet{wikipatretriever} leverages the graph structure induced by the entity links present in Wikipedia paragraphs to perform iterative multi-step retrieval. GoldEn (Gold Entity) retriever~\cite{golden} iteratively generates text queries based on paragraphs retrieved from an off-the-shelf retriever but requires training data for this next query generator. \citet{webgpt} used GPT3 to answer long-form questions by interacting with the browser but relied on human annotations of these interactions. All of these methods rely on supervised training on a large-scale dataset and can not be easily extended to a few-shot setting.

\section{Chain-of-Thought-Guided Retrieval and Open-Domain QA}
\label{sec:method}

\begin{figure*}[!ht]
\centering
\includegraphics[width=0.95\textwidth]{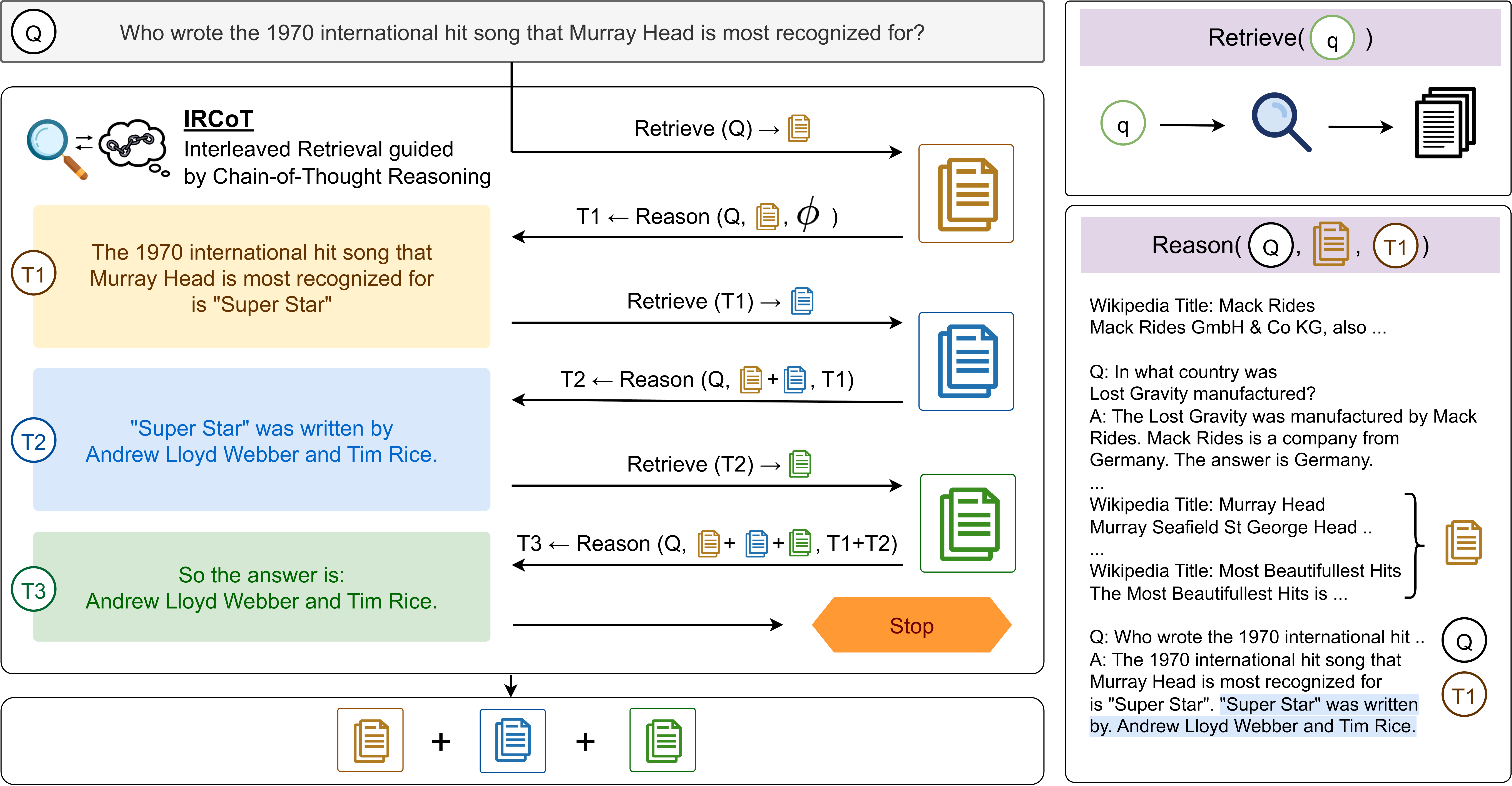}
\caption{\iconsys interleaves chain-of-thought (CoT) generation and retrieval steps to guide the retrieval by CoT and vice-versa. We start by retrieving $K$ documents using the question as they query and repeat two steps alternatingly until termination. (i) \texttt{reason}-step generates next CoT sentence based on the question, so far retrieved paragraphs, and CoT sentences. (ii) \texttt{retrieve}-step retrieves $K$ more paragraphs based on the last CoT sentence. The process terminates when the generated CoT has ``answer is'' or the number of steps exceeds a threshold. The collection of all paragraphs is returned as the retrieval result on the termination.}
\label{fig:main-figure}
\end{figure*}

Our goal is to answer a knowledge-intensive multi-step reasoning question $Q$ in a few-shot setting by using a knowledge source containing a large number of documents. To do this we follow a \texttt{retrieve-and-read} paradigm~\cite{retrieve-and-read}, where the retriever first retrieves documents from the knowledge source and the QA model reads the retrieved documents and the question to generate the final answer. Our contribution is mainly in the \texttt{retrieve} step (\S\ref{subsec:retriever}), and we use standard prompting strategies for the \texttt{read} step (\S\ref{subsec:reader}).

As noted earlier, for multi-step reasoning, retrieval can help guide the next reasoning step, which in turn can inform what to retrieve next. This motivates our interleaving strategy, discussed next.

\subsection{Interleaving Retrieval with Chain-of-Thought Reasoning}
\label{subsec:retriever}

Our proposed retriever method, \iconsys, can be instantiated from the following three ingredients: (i) a base retriever that can take a query and return a given number of paragraphs from a corpus or knowledge source; (ii) a language model with zero/few-shot Chain-of-Thought (CoT) generation capabilities; and (iii) a small number of annotated questions with reasoning steps explaining how to arrive at the answer in natural language (chain of thoughts) and a set of paragraphs from the knowledge source that collectively support the reasoning chain and the answer.

The overview of \iconsys is given in Fig.~\ref{fig:main-figure}. We first gather a base set of paragraphs by retrieving $K$ paragraphs using the question $Q$ as the query. Then, we interleave two steps (\texttt{reason} and \texttt{retrieve}) iteratively until the termination criterion is met.

The \textbf{retrieval-guided reasoning step (``Reason'')} generates the next CoT sentence using the question, the paragraphs collected thus far, and the CoT sentences generated thus far. The prompt template for the task looks as follows:

\begin{small}
\begin{verbatim}
Wikipedia Title: <Page Title>
<Paragraph Text>
...
Wikipedia Title: <Page Title>
<Paragraph Text>

Q: <Question>
A: <CoT-Sent-1> ... <CoT-Sent-n>
\end{verbatim}
\end{small}
 
For in-context demonstrations, we use the complete CoT in the above format. For a test instance, we show the model only the CoT sentences generated thus far and let it complete the rest. Even though the model may output multiple sentences, for each \texttt{reason-step}, we only take the first generated sentence and discard the rest.

For the paragraphs in the in-context demonstrations, we use ground-truth supporting paragraphs and $M$ randomly sampled paragraphs shuffled and concatenated together in the above format. For a test instance, we show all the paragraphs collected thus far across all the previous \texttt{retrieve-step}s.

If the generated CoT sentence has the ``answer is:'' string or the maximum number of steps\footnote{set to 8 in our experiments.} has been reached, we terminate the process and return all collected paragraphs as the retrieval result.

The \textbf{CoT-guided retrieval step (``Retrieve'')} uses the last generated CoT sentence as a query to retrieve more paragraphs and adds them to the collected paragraphs. We cap the total number of collected paragraphs\footnote{set to 15 in our experiments.} so as to fit in at least a few demonstrations in the model's context limit.

\subsection{Question Answering Reader}
\label{subsec:reader}

The QA reader answers the question using retrieved paragraphs taken from the retriever. We consider two versions of the QA reader implemented via two prompting strategies: \texttt{CoT Prompting} as proposed by \citet{cot}, Direct Prompting as proposed by \citet{originalgpt3}. For CoT prompting, we use the same template as shown in \S\ref{subsec:reader}, but at test time we ask the model to generate the full CoT from scratch. The final sentence of CoT is expected to be of the form ``answer is: ...'', so that the answer can be extracted programmatically. If it's not in that form, the full generation is returned as the answer. For Direct Prompting, we use the same template as CoT Prompting but the answer field (``A: '') contains only the final answer instead of CoT. See App.~\ref{sec:apndx-prompts} for details.

\section{Experimental Setup}
\label{sec:exp-setup}

We evaluate our method on 4 multi-step QA datasets in the open-domain setting: \textbf{HotpotQA}~\cite{hotpotqa}, \textbf{2WikiMultihopQA}~\cite{xanh2020_2wikimultihop}, answerable subset of \textbf{MuSiQue}~\cite{musique}, and answerable subset of \textbf{IIRC}~\cite{iirc}. For HotpotQA, we use the Wikipedia corpus that comes with it for the open-domain setting. For each of the other three datasets, which originally come in a reading comprehension or mixed setting, we used the associated contexts to construct a corpus for our open-domain setting (see App.~\ref{sec:apndx-corpora} for details). For each dataset, we use 100 randomly sampled questions from the original development set for tuning hyperparameters, and 500 other randomly sampled questions as our test set.

\subsection{Models}
\label{subsec:exp-models}

\paragraph{Retriever.} We use BM25~\cite{bm25} implemented in Elasticsearch\footnote{\url{https://www.elastic.co/}} as our base retriever. We compare two retriever systems:

(i) \textbf{One-step Retriever (OneR)} uses the question as a query to retrieve $K$ paragraphs. We select $K \in \{5, 7, 9, 11, 13, 15\}$ that's best on the dev set.

(ii) \textbf{\iconsys Retriever} is our method described in \S\ref{sec:method}. We use BM25 as its underlying retriever and experiment with OpenAI GPT3 (\texttt{code-davinci-002})~\cite{originalgpt3,instructgpt3,codex} and Flan-T5~\cite{flan} of different sizes as its CoT generator.

For demonstrating in-context examples to these LMs, we wrote CoTs for 20 questions for all the datasets (see App. \S\ref{sec:apndx-prompts}). We then create 3 demonstration (``training'') sets by sampling 15 questions each for each dataset. For each experiment, we search for the best hyperparameters for the dev set using the first demonstration set and evaluate each demonstration set on the test set using the selected hyperparameters. We report the mean and standard deviation of these 3 results for each experiment.

 At test time, we pack as many demonstrations as possible within the model's context length limit. The context limit for GPT3 (\texttt{code-davinci-002}) is 8K word pieces. Flan-T5-* doesn't have any hard limit as it uses relative position embeddings. But we limit Flan-T5's context to 6K word pieces, which is the maximum we could fit in the memory of our 80G A100 GPUs.

\iconsys Retriever has one key hyperparameter: $K \in \{2, 4, 6, 8\}$, the number of paragraphs to retrieve at each step. Additionally, when creating ``training'' demonstrations for \iconsys's Reasoner module, we use gold paragraphs and a smaller number $M \in \{1, 2, 3\}$ of distractor paragraphs (\S\ref{subsec:retriever}).

\textbf{Retrieval Metric:} We allow a maximum of 15 paragraphs for all retriever systems and measure the recall of the gold paragraphs among the retrieved set of paragraphs. We search for the hyperparameter $K$ (and $M$ for \iconsys) that maximizes the recall on the dev set and use it on the test set. The reported metric can thus be viewed as the \emph{fixed-budget optimal recall} for each system considered.\footnote{\label{footnote:retrieval-metric}Note that our retrieved documents are not ranked, making standard information retrieval metrics such as MAP and DCG inapplicable. Further, we can only limit the number of retrieved paragraphs \emph{per step} to $K$. Since the total number of reasoning steps varies for questions, and in some cases, we don't even obtain all $K$ paragraphs in a given step, the total number of retrieved paragraphs also varies (even though capped at 15). This makes Recall@k, Precision@k, etc., also not applicable as metrics for any given k.}

\paragraph{QA Reader.} To implement the reader, we use the same LMs as used in the \texttt{reason-step} of \iconsys Retriever. We found that QA readers implemented with Flan-T5-* perform better with the Direct Prompting strategy and GPT3 performs better with CoT Prompting strategy (see App. ~\ref{sec:apndx-readers-results}). Hence we use Direct prompting strategy for QA with Flan-T5-* and CoT with GPT3 for the experiments.\footnote{\iconsys, by construction, produces a CoT as a part of its retrieval process. Thus, instead of having a separate post-hoc reader, one can also just extract the answer from the CoT generated during retrieval. However, we found this to be a suboptimal choice, so we always use a separate reader (see App.~\ref{sec:qa-reader-ablation}).}

The QA reader has one hyperparameter $M$: the number of distractor paragraphs in the in-context demonstrations. We search for $M$ in $\{1, 2, 3\}$. When used in conjunction with \iconsys retriever $M$ is tied for the CoT generator and the reader.

\paragraph{Open-Domain QA (ODQA) Models.} Putting retrievers and readers together, we experiment with ODQA models constructed from the various language models denoted as \textbf{OneR QA} and \textbf{\iconsys QA}. For \iconsys QA, the choice of LM for the CoT generator and the reader is kept the same. We also experiment with retriever-less QA readers \textbf{NoR QA} to assess how well LMs can answer the question from their parametric knowledge alone. To select the best hyperparameters for the ODQA model, we search for the hyperparameters $K$ and $M$ that maximize the answer F1 on the development set.

IIRC is structured slightly differently from the other datasets, in that its questions are grounded in a main passage and other supporting paragraphs come from the Wikipedia pages of entities mentioned in this passage. We slightly modify the retrievers and readers to account for this (see App.~\ref{sec:apndx-iirc-special-handling}).

\begin{figure*}[ht]
\centering
\includegraphics[width=0.95\textwidth]{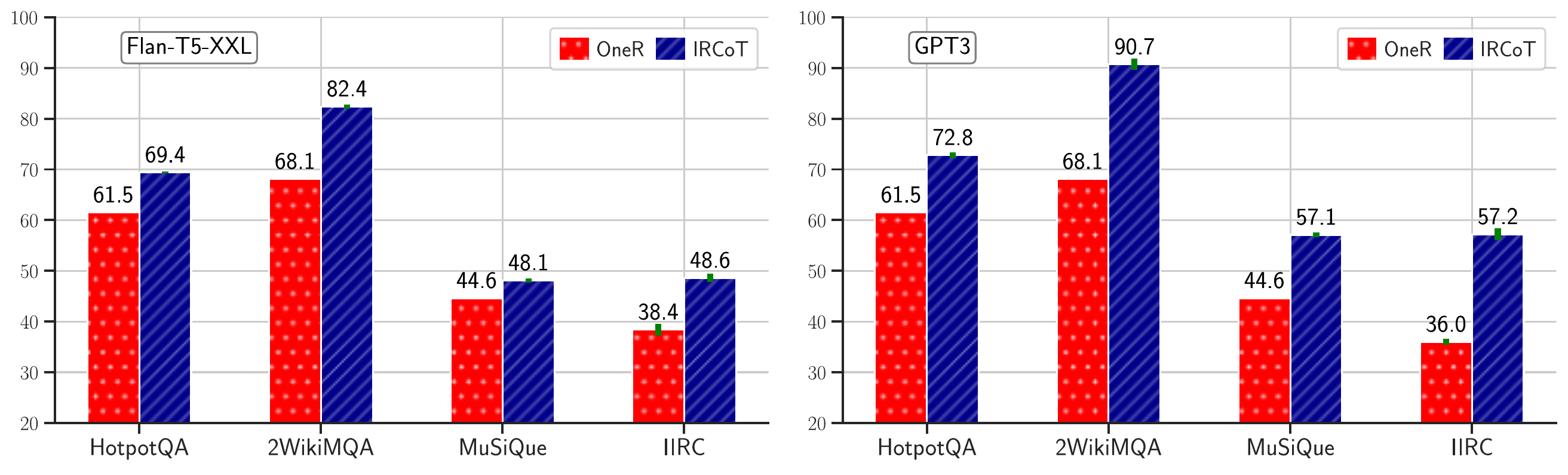}
\caption{Retrieval recall for one-step retriever (OneR) and \iconsys instantiated from \texttt{Flan-T5-XXL} (left) and \texttt{GPT3} (right) models. \iconsys outperforms OneR for both models and all datasets.}
\label{fig:main-retrieval-results}
\end{figure*}

\begin{figure*}[ht]
\centering
\includegraphics[width=0.95\textwidth]{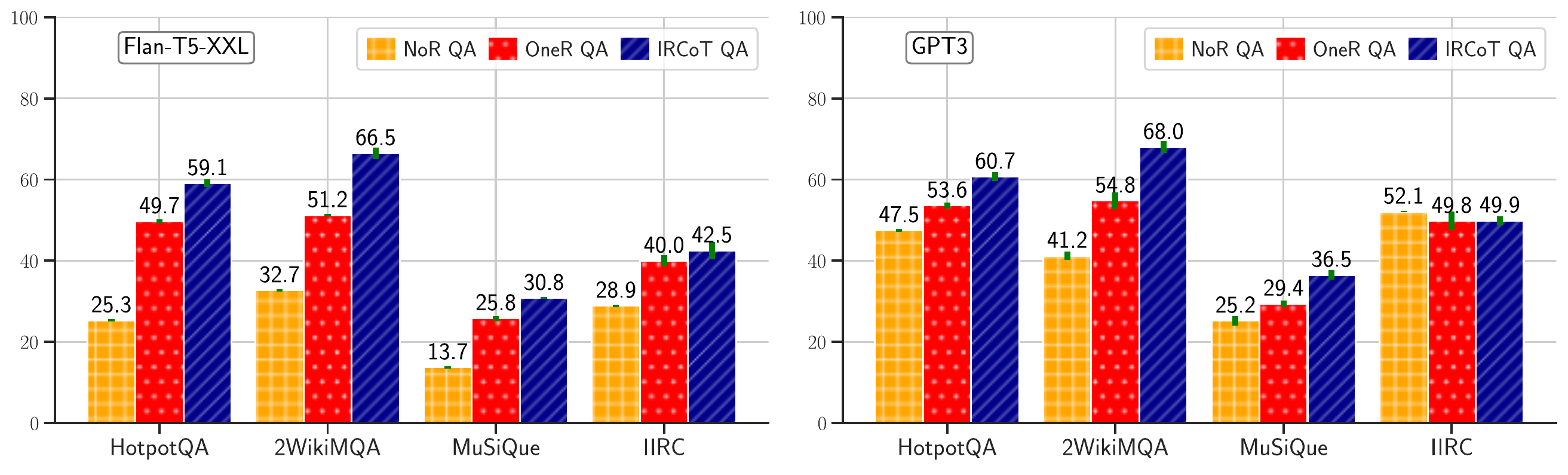}
\caption{Answer F1 for ODQA model made using (i) no retriever (NoR QA) (ii) one-step retriever (OneR QA) and (iii) \iconsys QA instantiated from \texttt{Flan-T5-XXL} (left) and \texttt{GPT3} (right) models. \iconsys QA outperforms OneR QA and NoR QA for both models on all datasets, except for GPT3 on IIRC.}
\label{fig:main-qa-results}
\end{figure*}
\section{Results}
\label{sec:exp-results}

\begin{figure*}[ht]
\centering
\includegraphics[width=0.95\textwidth]{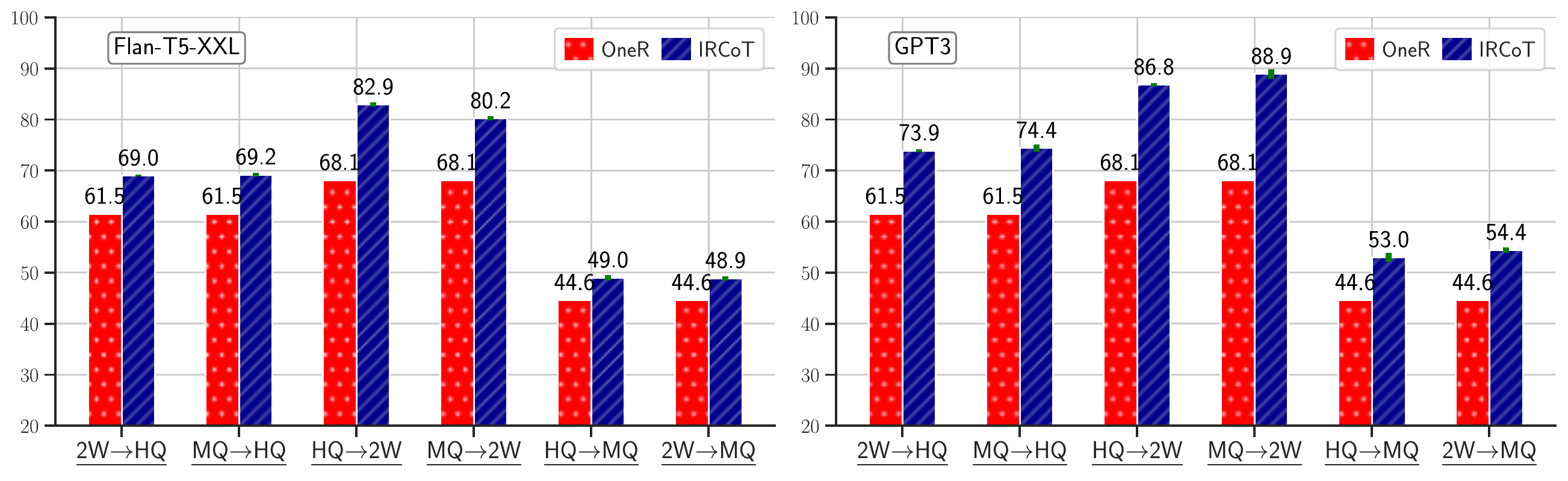}
\caption{Retrieval recall for OneR and IRCoT using Flan-T5-XXL (Left) and GPT3 (Right) in out-of-distribution (OOD) setting. HQ (HotpotQA), 2W (2WikiMultihopQA), MQ (MuSiQue). The result X$\rightarrow$Y indicates prompt demonstrations are from dataset X and evaluation is on dataset Y. \iconsys outperforms OneR in such an OOD setting.}
\label{fig:ood-retrieval-results}
\end{figure*}

\begin{figure*}[ht]
\centering
\includegraphics[width=0.95\textwidth]{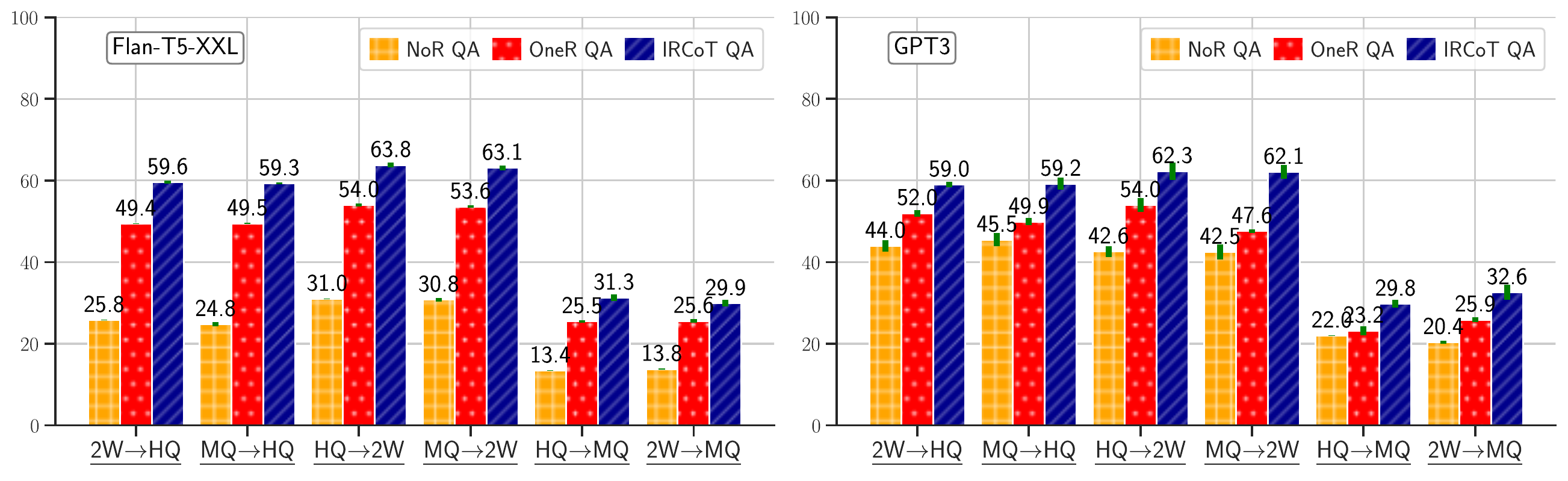}
\caption{Answer F1 for NoR QA, OneR QA and IRCoT QA using Flan-T5-XXL (Left) and GPT3 (Right) in out-of-distribution (OOD) setting. HQ (HotpotQA), 2W (2WikiMultihopQA), MQ (MuSiQue). The result X$\rightarrow$Y indicates prompt demonstrations are from dataset X and evaluation is on dataset Y. \iconsys QA outperforms OneR QA and NoR QA in such OOD setting.}
\label{fig:ood-qa-results}
\end{figure*}

\paragraph{\iconsys retrieval is better than one-step. }

Fig.~\ref{fig:main-retrieval-results} compares OneR with \iconsys retrievers made from \texttt{Flan-T5-XXL} and \texttt{GPT3} LMs. For both models, \iconsys significantly outperforms one-step retrieval across all datasets. For \texttt{Flan-T5-XXL}, \iconsys improves our recall metric relative to one-step retrieval, on HotpotQA by 7.9, on 2WikiMultihopQA by 14.3, on MuSiQue by 3.5, and on IIRC by 10.2 points. For \texttt{GPT3}, this improvement is by 11.3, 22.6, 12.5, and 21.2 points, respectively.

\paragraph{\iconsys QA outperforms NoR and OneR QA.}

Fig.~\ref{fig:main-qa-results} compares ODQA performance using NoR, OneR and \iconsys retriever made from \texttt{Flan-T5-XXL} and \texttt{GPT3} LMs. For \texttt{Flan-T5-XXL}, \iconsys QA outperforms OneR QA on HotpotQA by 9.4, on 2WikiMultihopQA by 15.3, on MuSiQue by 5.0 and IIRC by 2.5 F1 points. For \texttt{GPT3}, the corresponding numbers (except for IIRC) are 7.1, 13.2, and 7.1 F1 points. For \texttt{GPT3}, \iconsys doesn't improve the QA score on IIRC, despite significantly improved retrieval (21 points as shown in Fig.~\ref{fig:main-retrieval-results}). This is likely because IIRC relevant knowledge may already be present in GPT3, as also evidenced by its NoR QA score being similar. For other datasets and model combinations, NoR QA is much worse than \iconsys QA, indicating the limits of the models' parametric knowledge.

\begin{figure}[ht]
\centering
\includegraphics[width=0.475\textwidth]{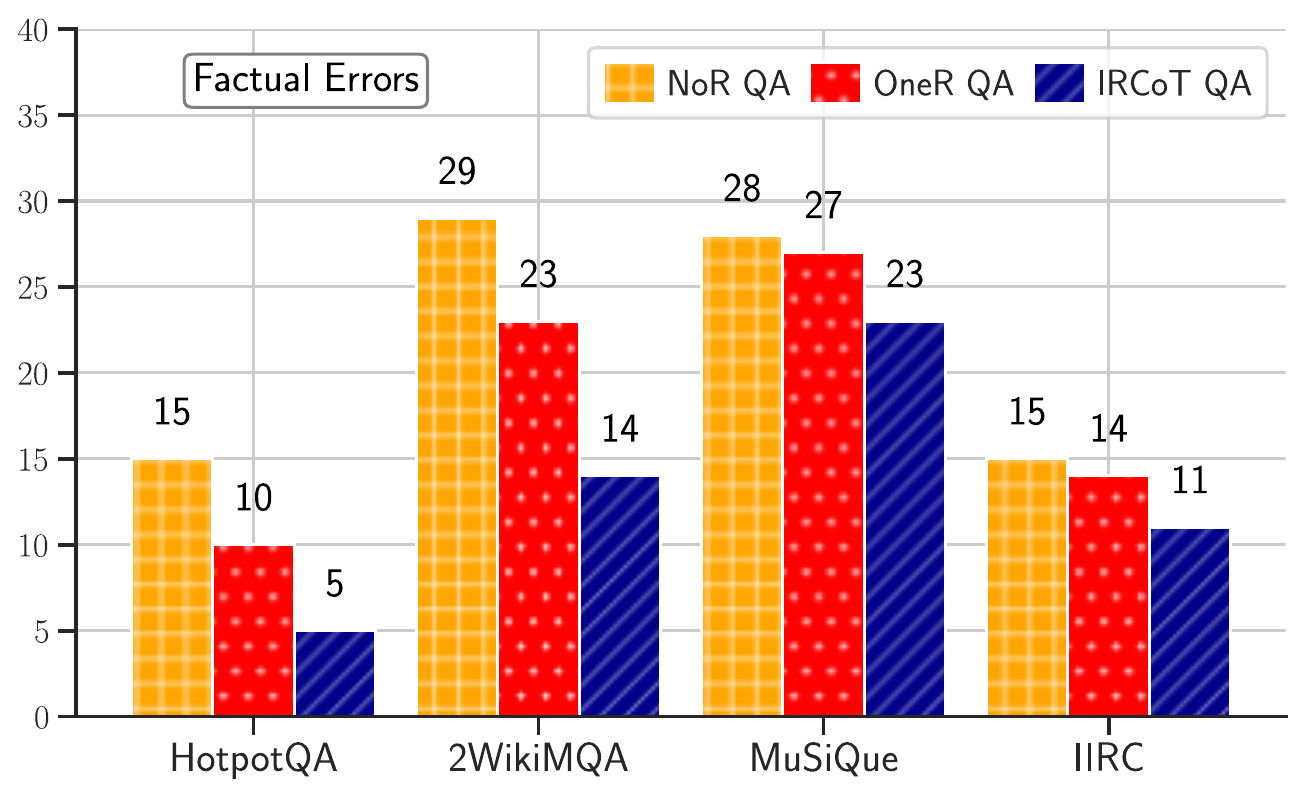}
\caption{Number of questions, out of 40, where CoT generated by GPT3 using different methods has at least 1 factual error. Factual errors: \iconsys $<$ OneR $<$ NoR.}
\label{fig:cot-factual-errors}
\end{figure}

\begin{figure*}[ht]
\centering
\includegraphics[width=0.95\textwidth]{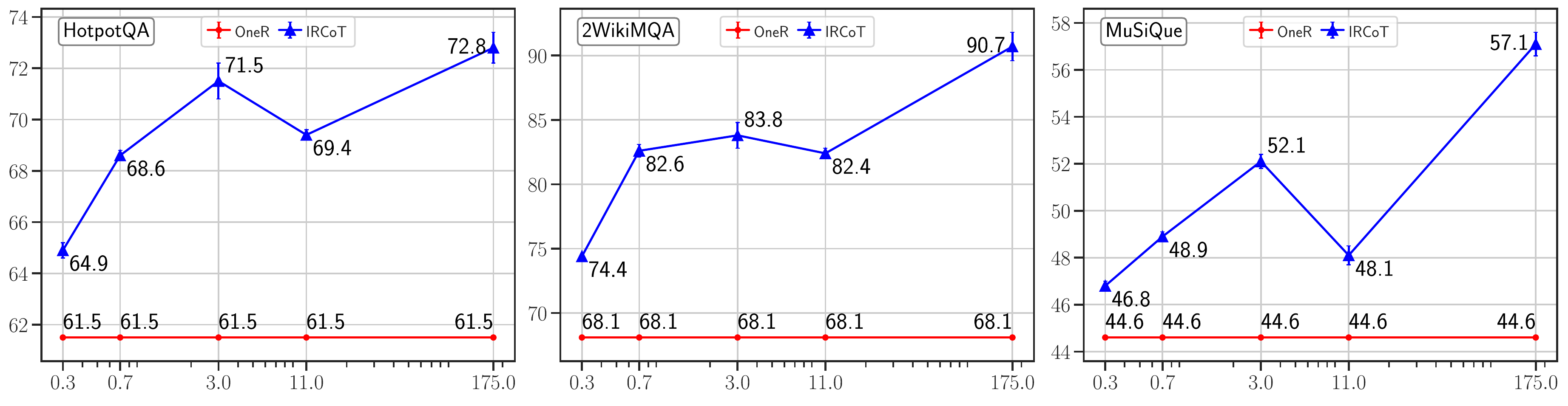}
\caption{Retrieval recall for OneR (bottom) and \iconsys (top) for LMs of increasing sizes: Flan-T5 \{base (0.2B), large (0.7B), XL (3B), XXL (11B)\} and GPT3 (175B) on HotpotQA, 2WikiMultihopQA, MuSiQue. \iconsys outperforms OneR for all model sizes, including the 0.3B model, and the difference roughly grows with model size. Note: OneR doesn't use LM in its retrieval and so has a fixed score.}
\label{fig:model-scale-retrieval-results}
\end{figure*}

\begin{figure*}[ht]
\centering
\includegraphics[width=0.95\textwidth]{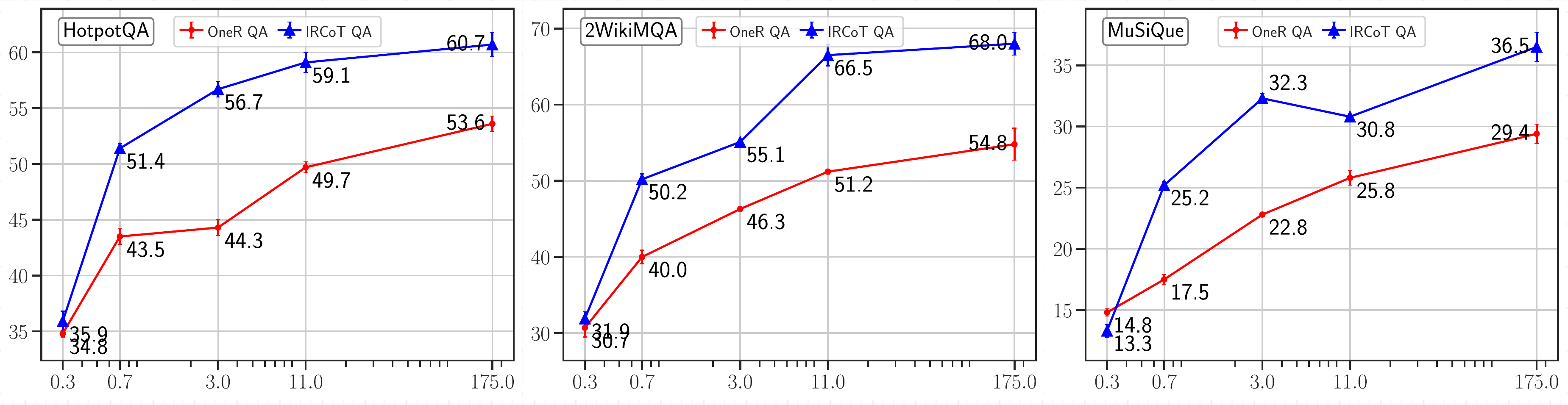}
\caption{Answer F1 for ODQA models made using OneR (bottom) and \iconsys (top) for LMs of increasing sizes: Flan-T5 \{base (0.2B), large (0.7B), XL (3B), XXL (11B)\} and GPT3 (175B) on HotpotQA, 2WikiMultihopQA and MuSiQue. \iconsys QA outperforms OneR QA for all model sizes except for the smallest, 0.3B. \iconsys with 3B model even outperforms OneR with 58X larger GPT3 model showing the value of improved retrieval.}
\label{fig:model-scale-qa-results}
\end{figure*}

\paragraph{\iconsys is effective in OOD setting. }

Since CoT may not always be easy to write for new datasets, we evaluate NoR, OneR, and IRCoT on generalization to new datasets, i.e. OOD setting. To do so, we use prompt demonstrations from one dataset to evaluate on another dataset.\footnote{We use the evaluation dataset's corpus for retrieval.} For all pairs of the datasets\footnote{We skip IIRC in this exploration as the task is structured a bit differently and requires special handling (see App.~\ref{sec:apndx-iirc-special-handling}).} and for both \texttt{Flan-T5-XXL} and \texttt{GPT3}, we find the same trend as in the IID setting: \iconsys retrieval outperforms OneR (Fig.~\ref{fig:ood-retrieval-results}), and IRCoT QA outperforms both OneR QA and NoR QA (Fig.~\ref{fig:ood-qa-results}).

\paragraph{\iconsys generates CoT with fewer factual errors.}

To assess whether our approach also improves the factuality of generated CoTs, we manually annotated CoTs generated by NoR QA, OneR QA, and IRCoT QA using GPT3 for 40 randomly sampled questions from each of the four datasets. We considered CoT to have a factual error if at least one of the facts\footnote{all sentences before the final ``answer is:'' sentence.} is not true.\footnote{Note that factual error doesn't necessarily mean the predicted answer is incorrect and vice-versa. This is because the model can generate a wrong answer despite all correct facts, and vice-versa. We also account for the possibility of answer annotation errors in the original datasets.} As Fig.~\ref{fig:cot-factual-errors} shows, NoR makes the most factual errors, OneR makes fewer, and \iconsys the least. In particular, \iconsys reduces the factual errors over OneR by 50\% on HotpotQA and 40\% on 2WikiMultihopQA.

Table~\ref{table:nor-oner-cot-examples} illustrates how the CoT predictions for different methods vary qualitatively. Since NoR relies completely on parametric knowledge, it often makes a factual error in the first sentence, which derails the full CoT. OneR can retrieve relevant information closest to the question and is less likely to make such errors early on, but it still makes errors later in the CoT. IRCoT, on the other hand, is often able to prevent such errors in each step.

\paragraph{\iconsys is also effective for smaller models.}

To see how effective \iconsys is at different LM sizes, we show the scaling plots in Fig.~\ref{fig:model-scale-retrieval-results}.\footnote{We skip IIRC here as the smaller models are not good at identifying Wikipedia titles from a paragraph and a question which is necessary for IIRC (see App.~\ref{sec:apndx-iirc-special-handling}).} We compare the recall for OneR and \iconsys using \texttt{Flan-T5} \{base (0.2B), large (0.7B), XL (3B), XXL (11B)\}, and GPT3 \texttt{code-davinci-002} (175B). \iconsys with even the smallest model (0.2B) is better than OneR, and the performance roughly improves with the model size. This shows the CoT generation capabilities of even small models can be leveraged for improving retrieval. Furthermore, we show the effect of model size on the QA score in Fig.~\ref{fig:model-scale-qa-results}. For all sizes except the smallest (0.2B), we see \iconsys QA is better than OneR QA. Moreover, \iconsys with a 3B model even outperforms OneR and NoR with a 58X larger 175B GPT3 model in all datasets.

\paragraph{\iconsys is SOTA for few-shot multistep ODQA.\footnote{\label{footnote:sota}App.~\S\ref{sec:sota-differences} reports updated SOTA numbers, including contemporaneous and newer works.}}

We compare \iconsys QA with five recent approaches to using LLMs for ODQA: Internet-Augmented QA~\cite{internet-augmented-qa}, RECITE~\cite{recitationlm} ReAct~\cite{react}, SelfAsk~\cite{selfask}, and DecomP~\cite{old-decomp}. Although these are not head-to-head comparisons as different methods use different APIs, knowledge sources, and even LLMs (see App.~\ref{sec:sota-differences} for details), it is still informative to explore, in a leaderboard-style fashion, how \iconsys performs relative to the best numbers published for these recent systems.

\vspace{0.1cm}
\begin{table}[ht]
    \centering
    \footnotesize
    \setlength{\tabcolsep}{2.0pt}
    \begin{tabular}{ccccc}\toprule
        Model &  HpQA\textsuperscript{Br}  &  HpQA & 2WikiMQA & MQ\textsuperscript{2H} \\
        \midrule
        InterAug      &   $-$ | $-$    &        30.3 | $-$\p{xx}   &        $-$ | $-$        &        $-$ | $-$            \\
        RECITE        &   $-$ | $-$    &        37.1 | 48.4        &        $-$ | $-$        &        $-$ | $-$            \\
        ReAct         &   $-$ | $-$    &        35.1 | $-$\p{xx}   &        $-$ | $-$        &        $-$ | $-$            \\
        SelfAsk       &   $-$ | $-$    &         $-$ | $-$         &       40.1 | $-$\p{xx}   &       15.2 | $-$\p{xx}     \\
        DecomP        &  \p{x..}$-$ | 50.0  &         $-$ | $-$         &   \p{x..}$-$ | 59.3      &       $-$ | $-$       \\
        \midrule
        \sys QA       &   \textbf{45.8 | 58.5}   &    \bf{49.3 | 60.7}       &   \bf{57.7 | 68.0}      &  \bf{34.2 | 43.8} \\
        \bottomrule
    \end{tabular}
    \caption{Comparison with other LLM-based ODQA systems on EM and F1 scores. `$-$': score is unavailable. HpQA\textsuperscript{Br}: Bridge questions subset of HotpotQA. MQ\textsuperscript{2H}: MuSiQue 2-hop questions. \iconsys QA with GPT3 (ours) outperforms other systems by a large margin. Note: Comparisons aren't head-to-head as discussed in the text. App.~\S\ref{sec:sota-differences} reports updated SOTA numbers, including contemporaneous and newer works.
    \label{table:extrinsic-comparison}
    }
\end{table}

As shown in Table~\ref{table:extrinsic-comparison}, \iconsys QA significantly outperforms all of these recent systems by a large margin, setting a new state of the art in terms of what's achievable via retrieval-augmented LLMs (without supervised training).

\begin{table*}[t]
    \centering
    \footnotesize
    \setlength{\tabcolsep}{4.0pt}
    \begin{tabular}{llllll}\toprule
        & \textbf{Question} &   \textbf{Gold Facts} & \textbf{NoR QA} & \textbf{OneR QA} & \textbf{IRCoT QA} \\
        \midrule

        \multirow{2}{*}{\rotatebox[origin=c]{90}{\parbox[c]{3.0cm}{\centering HotpotQA}}}
        & \begin{minipage}[t]{2cm} The tap dancing soloist in My Feet Keep Dancing also starred in what 1935 MGM musical? \end{minipage}
        & \begin{minipage}[t]{3.0cm} $\Rightarrow$ The tap dancing soloist in My Feet Keep Dancing is \textcolor{blue}{Fayard Nicholas}. \\ $\Rightarrow$ Fayard Nicholas starred in the 1935 MGM musical \textcolor{blue}{An All-Colored Vaudeville Show}. \end{minipage}
        & \begin{minipage}[t]{3.0cm} The tap dancing soloist in My Feet Keep Dancing is \textcolor{red}{Eleanor Powell}. Eleanor Powell also starred in the 1935 MGM musical \textcolor{red}{Broadway Melody} of 1936. So the answer is: Broadway Melody of 1936. \end{minipage}
        & \begin{minipage}[t]{3.0cm} The tap dancing soloist in My Feet Keep Dancing is \textcolor{blue}{Fayard Nicholas}. Fayard Nicholas also starred in the 1935 MGM musical \textcolor{red}{Top Hat}. So the answer is: Top Hat. \end{minipage}
        & \begin{minipage}[t]{3.0cm} The tap dancing soloist in My Feet Keep Dancing is \textcolor{blue}{Fayard Nicholas}. Fayard Nicholas starred in the 1935 MGM musical \textcolor{blue}{An All-Colored Vaudeville Show}. So the answer is: An All-Colored Vaudeville Show. \end{minipage} \\

        \midrule

        \multirow{1}{*}{\rotatebox[origin=c]{90}{\parbox[c]{4.0cm}{\centering 2WikiMQA}}}
        & \begin{minipage}[t]{2cm} Who is the father-in-law of Queen Hyojeong? \end{minipage}
        & \begin{minipage}[t]{3.0cm} $\Rightarrow$ Queen Hyojeong is the wife of \textcolor{blue}{King Heonjong of Joseon}. \\ $\Rightarrow$ King Heonjong of Joseon is the son of \textcolor{blue}{Crown Prince Hyomyeong}. \end{minipage}
        & \begin{minipage}[t]{3.0cm} Queen Hyojeong has a husband named \textcolor{red}{King Gojong}. King Gojong has a father named \textcolor{red}{Heungseon Daewongun}. Thus, Queen Hyojeong has a father-in-law named Heungseon Daewongun. So the answer is: Heungseon Daewongun. \end{minipage}
        & \begin{minipage}[t]{3.0cm} Queen Hyojeong is the wife of \textcolor{blue}{King Heonjong of Joseon}. King Heonjong of Joseon is the son of \textcolor{red}{King Sejo of Joseon}. Thus, King Sejo of Joseon is the father-in-law of Queen Hyojeong. So the answer is: King Sejo of Joseon. \end{minipage}
        & \begin{minipage}[t]{3.0cm} Queen Hyojeong is the wife of \textcolor{blue}{King Heonjong of Joseon}. King Heonjong of Joseon is the son of \textcolor{blue}{Crown Prince Hyomyeong}. Thus, Crown Prince Hyomyeong is the father-in-law of Queen Hyojeong. So the answer is: Crown Prince Hyomyeong. \end{minipage} \\

        \midrule

        \multirow{1}{*}{\rotatebox[origin=c]{90}{\parbox[c]{3.0cm}{\centering MuSiQue}}}
        & \begin{minipage}[t]{2cm} What is the name of the castle in the city where the performer of A Collection 1984–1989 was born? \end{minipage}
        & \begin{minipage}[t]{3.0cm} $\Rightarrow$ A Collection 1984–1989 was performed by \textcolor{blue}{Jane Siberry}. \\ $\Rightarrow$ Jane Siberry was born in \textcolor{blue}{Toronto}. \\ $\Rightarrow$ The castle in Toronto is the \textcolor{blue}{Casa Loma}. \end{minipage}
        & \begin{minipage}[t]{3.0cm} The performer of A Collection 1984–1989 is \textcolor{red}{The The}. The The was born in \textcolor{red}{London}. The name of the castle in London is the \textcolor{red}{Tower of London}. So the answer is: the Tower of London. \end{minipage}
        & \begin{minipage}[t]{3.0cm} A Collection 1984–1989 was performed by \textcolor{blue}{Jane Siberry}. Jane Siberry was born in \textcolor{blue}{Toronto}. The castle in Toronto is \textcolor{red}{Peqin Castle}. So the answer is: Peqin Castle. \end{minipage}
        & \begin{minipage}[t]{3.0cm} A Collection 1984–1989 was performed by \textcolor{blue}{Jane Siberry}. Jane Siberry was born in \textcolor{blue}{Toronto}. The castle in Toronto is the \textcolor{blue}{Casa Loma}. So the answer is: Casa Loma. \end{minipage} \\

        \bottomrule

    \end{tabular}
    \caption{Example CoTs generated by GPT3 with different methods. Since NoR relies on parametric knowledge, it often makes a factual error in the first sentence derailing the full CoT. OneR can retrieve relevant information closest to the question and is less likely to make such errors early on, but it still makes errors later in the CoT. As \iconsys performs retrieval after each step, it is often able to prevent such errors in each step. More examples are in App.~\ref{sec:apdx-nor-oner-cot-examples}.}
    \label{table:nor-oner-cot-examples}
\end{table*}

\section{Conclusions}

Chain-of-thought prompting has significantly improved LLMs' ability to perform multi-step reasoning. We leveraged this ability to improve retrieval, and in turn, improve QA performance for complex knowledge-intensive open-domain tasks in a few-shot setting. We argued that one-step question-based retrieval is insufficient for such tasks, and introduced \iconsys, which uses interleaved CoT reasoning and retrieval steps that guide each other step-by-step. On four datasets, \iconsys significantly improves both retrieval and QA performance when compared to one-step retrieval, for both large and relatively smaller-scale LMs. Additionally, CoTs generated by \iconsys contain fewer factual errors.

\section*{Limitations}
\label{sec:limitations}

\iconsys relies on the base LM to have a zero or few-shot CoT-generation ability.  While this is commonly available in large LMs (over 100B), it's not as common for small LMs (under 20B), which to some extent limits \iconsys adoptability. Given the recent surge of interest~\cite{ul2,reasoningdistillation1,reasoningdistillation2}, however, smaller LMs will likely increasingly acquire such ability, making IRCoT compatible with many more LMs.

\iconsys also relies on the base LM to support long inputs as multiple retrieved paragraphs need to fit in the LM's input, in addition to at least a few demonstrations of QA or CoT with paragraphs. This was supported by the models we used as \texttt{code-davinci-002} (GPT3) allows 8K tokens and Flan-T5-* uses relative position embeddings making it as extensible as the GPU memory constraints allow. Future work can explore strategies to rerank and select the retrieved paragraphs instead of passing all of them to the LM to alleviate the need for the LM to support long input.

The performance gain of \iconsys retriever and QA (over OneR and ZeroR baselines) come with an additional computational cost. This is because \iconsys makes a separate call to an (L)LM for each sentence of CoT. Future work can focus on, for instance, dynamically deciding when to retrieve more information and when to perform additional reasoning with the current information.

Lastly, a portion of our experiments was carried out using a commercial LLM API from OpenAI (\texttt{code-davinci-002}). This model was deprecated by OpenAI after our submission making the reproduction of these experiments challenging despite our best efforts, just like any other work using such APIs. The trends discussed in the paper (\iconsys $>$ OneR $>$ NoR), we believe, would still hold. Additionally, all our experiments using Flan-T5-*, which exhibit similar trends as that of GPT3, will remain reproducible, thanks to its publicly available model weights.
\section*{Ethical Considerations}

Language models are known to hallucinate incorrect and potentially biased information. This is especially problematic when the questions asked to it are of a sensitive nature. While retrieval-augmented approaches such as ours are expected to alleviate this issue to some extent by grounding generation in external text, this by no means solves the problem of generating biased or offensive statements. Appropriate care should thus be taken if deploying such systems in user-facing applications.

All the datasets and models used in this work are publicly available with permissible licenses. HotpotQA has CC BY-SA 4.0 license\footnote{\url{https://creativecommons.org/licenses/by-sa/4.0/}}, 2WikiMultihopQA has Apache-2.0 license\footnote{\url{https://www.apache.org/licenses/LICENSE-2.0}}, MuSiQUe and IIRC have CC BY 4.0 license\footnote{\url{https://creativecommons.org/licenses/by/4.0}}, and Flan-T5-* models have Apache-2.0 license.
\section*{Acknowledgments}

We thank the reviewers for their valuable feedback and suggestions. We also thank OpenAI for providing access to the \texttt{code-davinci-002} API. This material is based on research supported in part by the Air Force Research Laboratory (AFRL), DARPA, for the KAIROS program under agreement number FA8750-19-2-1003, in part by the National Science Foundation under the award IIS \#2007290, and in part by an award from the Stony Brook Trustees Faculty Awards Program.

\bibliography{custom}
\bibliographystyle{acl_natbib}

\appendix
\clearpage
\appendix

\section{Constructing Retrieval Corpora}
\label{sec:apndx-corpora}

HotpotQA already comes with the associated Wikipedia corpus for the open-domain setting, so we use it directly. 2WikiMultihopQA and MuSiQue, however, are originally reading comprehension datasets. Questions in 2WikiMultihopQA and MuSiQue are associated with 10 and 20 paragraphs respectively, 2-4 of which are supporting and others are non-supporting. To turn these datasets into an open-domain setting, we make two corpora, one for each dataset, by combining all supporting and non-supporting paragraphs for all its questions in the train, development, and test sets. IIRC is originally a mix between reading comprehension and an open-domain setting. Each question is grounded in one main paragraph, which contains links to multiple Wikipedia pages with several paragraphs each. We create a corpus out of all the paragraphs from all the Wikipedia pages present in the dataset.\footnote{Following are the corpus sizes for the datasets: HotpotQA (5,233,329), 2WikiMultihopQA (430,225), MuSiQue (139,416), and IIRC (1,882,415)} We do assume the availability of the main passage which doesn't need to be retrieved and is always present. We don't assume the availability of Wikipedia links in the main passage, however, to keep the retrieval problem challenging.\footnote{IIRC corpus has a positional bias, i.e., the majority of supporting paragraphs are always within the first few positions of the Wikipedia page. To keep the retrieval problem challenging enough we shuffle the paragraphs before indexing the corpus, i.e., we don't use positional information in any way.}

\section{Special Handling of Models for IIRC}
\label{sec:apndx-iirc-special-handling}

IIRC is slightly different from the other datasets, in that the question is grounded in the main passage and other supporting paragraphs come from the Wikipedia pages of entities mentioned in this passage. We modify the retrievers and readers to account for this difference: (i) We always keep the main passage as part of the input to the model regardless of the retrieval strategy used. (ii) For all the retrieval methods, we first prompt the model to generate a list of Wikipedia page titles using the main passage and the question. We map these generated titles to the nearest Wikipedia page titles in the corpus (found using BM25), and then the rest of the paragraph retrieval queries are scoped within only those Wikipedia pages.

To prompt the model to generate Wikipedia page titles using the main passage and the question for IIRC, we use the following template.

\begin{small}
\begin{verbatim}
Wikipedia Title: <Main Page Title>
<Main Paragraph Text>

Q: The question is: '<Question>'. Generate titles 
of <N> Wikipedia pages that have relevant
information to answer this question.
A: ["<Title-1>", "<Title-2>", ...]
\end{verbatim}
\end{small}

For ``training'', i.e., for demonstrations, N ($\le 3$) is the number of supporting Wikipedia page titles for the question. At test time, since the number of supporting page titles is unknown, we use a fixed value of 3. We found this trick of prompting the model to generate more titles at the test time improves its recall over letting the model decide by itself how many titles to generate.

\begin{table*}[ht]
    \centering
    \footnotesize
    \setlength{\tabcolsep}{10.0pt}
    \begin{tabular}{lccccc}\toprule
        Model &  HpQA\textsuperscript{Br} &  HpQA & 2WikiMQA & MQ\textsuperscript{2H} & MQ \\
        \midrule
        InterAug~\cite{internet-augmented-qa}  &         $-$ | $-$   &        30.3 | $-$\p{xx}   &        $-$ | $-$        &        $-$ | $-$        &   $-$ | $-$       \\
        RECITE~\cite{recitationlm}             &         $-$ | $-$   &        37.1 | 48.4        &        $-$ | $-$        &        $-$ | $-$        &   $-$ | $-$       \\
        ReAct~\cite{react}                     &         $-$ | $-$   &        35.1 | $-$\p{xx}   &        $-$ | $-$        &        $-$ | $-$        &   $-$ | $-$       \\
        SelfAsk~\cite{selfask}                 &         $-$ | $-$   &         $-$ | $-$         &       40.1 | $-$\p{xx}   &       15.2 | $-$\p{xx} &   $-$ | $-$        \\
        DecomP~\cite{old-decomp}               &  \p{x..}$-$ | 50.0  &         $-$ | $-$         &   \p{x..}$-$ | 59.3      &       $-$ | $-$        &   $-$ | $-$ \\
        \midrule
        DecomP~\cite{decomp} *                 &         $-$ | $-$   &         \p{x..}$-$ | 53.5 &   \p{x..}$-$ | \textbf{70.8}      &       $-$ | $-$        &  \p{xx}$-$ | 30.9 \\
        DSP~\cite{dsp} *                       &         $-$ | $-$   &     \bf{51.4 | 62.9}      &        $-$ | $-$        &        $-$ | $-$        &        $-$ | $-$   \\
        \midrule
        \sys QA (ours)                         & \textbf{45.8 | 58.5} &       49.3 | 60.7       &   57.7 | 68.0      &  \bf{34.2 | 43.8}       &   \textbf{26.5 | 36.5}      \\
        \bottomrule
    \end{tabular}
    \caption{Extended comparison with published LLM-based ODQA systems (as of May 25, 2023) on EM and F1 scores (with new numbers marked with *). `$-$': score is unavailable. HpQA\textsuperscript{Br}: Bridge questions subset of HotpotQA. MQ\textsuperscript{2H}: MuSiQue 2-hop questions. IRCoT remains SOTA for MuSiQue and is close to SOTA for HotpotQA and 2WikiMultihopQA. Note the comparisons here are not head-to-head as discussed in the text.}
    \label{table:extended-extrinsic-comparison}
\end{table*}

\begin{table*}[ht]
    \centering
    \footnotesize
    \setlength{\tabcolsep}{3.5pt}
    \begin{tabular}{cccccccccccc}\toprule
        & &\p{0}& \multicolumn{4}{c}{Flan-T5-XXL} & \p{0}& \multicolumn{4}{c}{GPT3} \\
        \cmidrule{4-7} \cmidrule{9-12}
        & Model &\p{0}&   HotpotQA & 2WikiMQA & MuSiQue & IIRC &  \p{0}&   HotpotQA & 2WikiMQA & MuSiQue & IIRC\\
        \midrule
        \multirow{2}{*}{ZeroR QA}
        & Direct       &\p{0}&        \bf{25.3}\std{0.3}  &  \bf{32.7}\std{0.3}  &  \bf{13.7}\std{0.3}  &  \bf{28.9}\std{0.3} &   \p{0}&     \nf{41.0}\std{1.1}  &  \nf{38.5}\std{1.1}  &  \nf{19.0}\std{1.2} &  \nf{40.9}\std{0.7} \\
        & CoT          &\p{0}&        \nf{22.9}\std{0.1}  &  \nf{31.7}\std{1.5}  &  \nf{10.3}\std{0.5}  &  \nf{24.4}\std{0.1} &   \p{0}&     \bf{47.5}\std{0.4}  &  \bf{41.2}\std{1.0}  &  \bf{25.2}\std{1.2} &  \bf{52.1}\std{0.1} \\
        \midrule
        \multirow{2}{*}{OneR QA}
        & Direct       &\p{0}&        \bf{49.7}\std{0.5}  &  \bf{51.2}\std{0.3}  &  \bf{25.8}\std{0.6}  &  \bf{40.0}\std{1.3} &   \p{0}&     \nf{50.7}\std{0.1}  &  \nf{46.4}\std{2.9}  &  \nf{20.4}\std{0.3} &  \nf{40.1}\std{0.9} \\
        & CoT          &\p{0}&        \nf{43.1}\std{0.7}  &  \nf{47.8}\std{0.9}  &  \nf{17.6}\std{0.2}  &  \nf{34.5}\std{1.5} &   \p{0}&     \bf{53.6}\std{0.7}  &  \bf{54.8}\std{2.1}  &  \bf{29.4}\std{0.8} &  \bf{49.8}\std{2.3} \\
        \midrule
        \multirow{2}{*}{\fixedicon\sys QA}
        & Direct       &\p{0}&        \bf{59.1}\std{0.9}  &  \bf{66.5}\std{1.4}  &  \bf{30.8}\std{0.2}  &  \bf{42.5}\std{2.1} &   \p{0}&     \nf{60.6}\std{1.0}  &  \nf{63.5}\std{2.7}  &  \nf{36.0}\std{0.5} &  \nf{47.9}\std{2.3} \\
        & CoT          &\p{0}&        \nf{52.0}\std{0.6}  &  \nf{55.1}\std{1.0}  &  \nf{24.9}\std{1.0}  &  \nf{36.5}\std{1.3} &   \p{0}&     \bf{60.7}\std{1.1}  &  \bf{68.0}\std{1.5}  &  \bf{36.5}\std{1.2} &  \bf{49.9}\std{1.1} \\
        \bottomrule
    \end{tabular}
    \caption{Answer F1 for different ODQA models made from NoR, One and \iconsys retrievals, and Direct and CoT prompting readers. For \texttt{Flan-T5-XXL}, Direct prompting is a better choice for the reader, and for \texttt{GPT3}, CoT prompting is a better choice for the reader. Hence, we make different reader choices for Flan-T5 and GPT3 for the experiments in the main paper. Note that \iconsys QA $>$ OneR QA $>$ ZeroR QA holds up regardless of this choice.}
    \label{table:direct-cot-reader-results}
\end{table*}

\begin{table*}[htb]
    \centering
    \footnotesize
    \setlength{\tabcolsep}{4.0pt}
    \begin{tabular}{llllll}\toprule
        & \textbf{Question} &   \textbf{Gold Facts} & \textbf{NoR QA} & \textbf{OneR QA} & \textbf{IRCoT QA} \\

        \midrule

        \multirow{2}{*}{\rotatebox[origin=c]{90}{\parbox[c]{2.5cm}{\centering HotpotQA}}}
        & \begin{minipage}[t]{2cm} What was the island, on which Marinelli Glacier is located, formerly known as?\end{minipage}
        & \begin{minipage}[t]{3.0cm} $\Rightarrow$ Marinelli Glacier is located on the island of \textcolor{blue}{Tierra del Fuego}. \\ $\Rightarrow$ The island of Tierra del Fuego was formerly known as \textcolor{blue}{Isla de Xativa}. \end{minipage}
        & \begin{minipage}[t]{3.0cm} Marinelli Glacier is located on the island of \textcolor{red}{Graham Land}. Graham Land was formerly known as \textcolor{red}{Graham's Land}. So the answer is: Graham's Land. \end{minipage}
        & \begin{minipage}[t]{3.0cm} Marinelli Glacier is located on \textcolor{blue}{Isla Grande de Tierra del Fuego}. Isla Grande de Tierra del Fuego was formerly known as \textcolor{red}{Tierra del Fuego}. So the answer is: Tierra del Fuego. \end{minipage}
        & \begin{minipage}[t]{3.0cm} Marinelli Glacier is located on the island of \textcolor{blue}{Tierra del Fuego}. The island of Tierra del Fuego was formerly known as \textcolor{blue}{Isla de Xativa}. So the answer is: Isla de Xativa. \end{minipage} \\

        \midrule

        \multirow{2}{*}{\rotatebox[origin=c]{90}{\parbox[c]{3.0cm}{\centering 2WikiMQA}}}
        & \begin{minipage}[t]{2cm} Who is the child of the director of film Mukhyamantri (1996 Film)? \end{minipage}
        & \begin{minipage}[t]{3.0cm} $\Rightarrow$ The director of Mukhyamantri (1996 film) is \textcolor{blue}{Anjan Choudhury}. \\ $\Rightarrow$ Anjan Choudhury has a child named \textcolor{blue}{Sandip Choudhury}. \end{minipage}
        & \begin{minipage}[t]{3.0cm} The film Mukhyamantri was directed by \textcolor{red}{S. V. Rajendra Singh Babu}. S. V. Rajendra Singh Babu has a child named \textcolor{red}{S. V. Rajendra Singh Babu}. So the answer is: S. V. Rajendra Singh Babu. \end{minipage}
        & \begin{minipage}[t]{3.0cm} Mukhyamantri (1996 film) was directed by \textcolor{blue}{Anjan Choudhury}. Anjan Choudhury has a child named \textcolor{red}{Aniruddha Choudhury}. So the answer is: Aniruddha Choudhury. \end{minipage}
        & \begin{minipage}[t]{3.0cm} The director of Mukhyamantri (1996 film) is \textcolor{blue}{Anjan Choudhury}. Anjan Choudhury has a child named \textcolor{blue}{Sandip Choudhury}. So the answer is: Sandip Choudhury. \end{minipage} \\

        \midrule

        \multirow{2}{*}{\rotatebox[origin=c]{90}{\parbox[c]{2.7cm}{\centering MuSiQue}}}
        & \begin{minipage}[t]{2cm} When did the Governorship end of the city where the author of Sacerdotii Nostri Primordia died? \end{minipage}
        & \begin{minipage}[t]{3.0cm} $\Rightarrow$ Sacerdotii Nostri Primordia was written by \textcolor{blue}{Pope John XXIII}. \\ $\Rightarrow$ Pope John XXIII died in \textcolor{blue}{Vatican City}. \\ $\Rightarrow$ The Governorship of Vatican City ended in \textcolor{blue}{1952}. \end{minipage}
        & \begin{minipage}[t]{3.0cm} The author of Sacerdotii Nostri Primordia is \textcolor{red}{Pope Pius IX}. Pope Pius IX died in the city of \textcolor{red}{Rome}. The Governorship of Rome ended in \textcolor{red}{1870}. So the answer is: 1870. \end{minipage}
        & \begin{minipage}[t]{3.0cm} Sacerdotii Nostri Primordia was written by \textcolor{blue}{Pope John XXIII}. Pope John XXIII died in the city of \textcolor{red}{Rome}. The Governorship of Rome ended in \textcolor{red}{1870}. So the answer is: 1870. \end{minipage}
        & \begin{minipage}[t]{3.0cm} Sacerdotii Nostri Primordia was written by \textcolor{blue}{Pope John XXIII}. Pope John XXIII died in \textcolor{blue}{Vatican City}. The Governorship of Vatican City ended in \textcolor{blue}{1952}. So the answer is: 1952. \end{minipage} \\

        \bottomrule

    \end{tabular}
    \caption{Additional CoTs generated by GPT3 with different methods. ZeroR is most prone to factual errors. OneR often fixes some of the factual information which is closest to the question but doesn't always fix it all the way. Since IRCoT retrieves after each step, it can also fix the errors at each step.  More examples are in Table~\ref{table:nor-oner-cot-examples}.}
    \label{table:apdx-nor-oner-cot-examples}
\end{table*}

\section{Comparison with Previous Systems for ODQA with LLMs}
\label{sec:sota-differences}

We showed a leaderboard-style comparison with previous approaches to using large language models for open-domain QA in \S~\ref{sec:exp-results}. We noted though that the comparison is not head-to-head given various differences. We briefly describe each method and the differences in API, LLM, retrieval corpus, and other choices here.

Internet-Augmented QA~\cite{internet-augmented-qa} does (one-step) Google Search retrieval, performs additional LLM-based filtering on it, and then prompts an LLM to answer the question using the resulting context. It uses the Gopher 280B language model. RECITE~\cite{recitationlm} bypasses the retrieval and instead prompts an LLM to first generate (recite) one or several relevant passages from its own memory, and generate the answer conditioned on this generation. They experiment with many LLMs, the highest performing of which is \texttt{code-davinci-002} which we report here. ReAct~\cite{react} prompts LLMs to produce reasoning and action traces where actions are calls to a Wikipedia API to return the summary for a given Wikipedia page title. It uses the PALM 540B model. SelfAsk~\cite{selfask} prompts LLMs to decompose a question into subquestions and answers these subquestions by issuing separate calls to the Google Search API. It uses the GPT3 (\texttt{text-davinci-002}) model. Finally, DecomP~\cite{decomp} is a general framework that decomposes a task and delegates sub-tasks to appropriate sub-models. Similar to our system, it uses BM25 Search and the GPT3 (\texttt{code-davinci-002}) model. And lastly, DSP~\cite{dsp} provides a way to programmatically define interactions between LLM and retrieval for ODQA (e.g., via question decomposition), bootstrap demonstrations for such a program, and use them to make the answer prediction. It uses GPT3.5 LLM with ColBERT-based retrieval. Since most of these methods use different knowledge sources or APIs and are built using different LLMs and retrieval models, it's difficult to make a fair scientific comparison across these systems. Additionally, the evaluations in the respective papers are on different random subsets (from the same distribution) of test instances. 

Despite these differences, it is still informative to explore, in a leaderboard-style fashion, how \iconsys performs relative to the best numbers published for these recent systems. Table~\ref{table:extended-extrinsic-comparison} shows results from different systems, including contemporaneous and newer numbers. The two new systems in this table (relative to Table~\ref{table:extrinsic-comparison}) are DecomP (newer version) and DSP. While \iconsys remains SOTA on MuSiQue, DSP outperforms it on HotpotQA by 2.0 points and the newer version of Decomp outperforms \iconsys on 2WikiMultihopQA by 2.8 points. We speculate DecomP performs well on 2WikiMultihopQA because it has only a few easy-to-predict decomposition patterns, which DecomP's question decomposition can leverage. The lack of such patterns in HotpotQA and MuSiQue causes it to underperform compared to \iconsys. Lastly, it will be useful to assess whether DSP, which is hardcoded for 2-hop questions like that of HotpotQA, will work well for a dataset with a varied number of hops like that of MuSiQue. We leave this further investigation to future work.

\section{Additional CoT Generation Examples}
\label{sec:apdx-nor-oner-cot-examples}

Table~\ref{table:apdx-nor-oner-cot-examples} provides illustrations, in addition to the ones provided in Table~\ref{table:nor-oner-cot-examples}, for how the CoT generations for NoR QA, OneR QA, and IRCoT QA methods vary. This gives an insight into how IRCoT improves QA performance. Since NoR relies completely on parametric knowledge, it often makes a factual error in the first sentence, which derails the full reasoning chain. Some of this factual information can be fixed by OneR, especially information closest to the question (i.e., can be retrieved using the question). This is insufficient for fixing all the mistakes. Since IRCoT involves retrieval after each step, it can fix errors at each step.

\section{Direct vs CoT Prompting Readers}
\label{sec:apndx-readers-results}

Table~\ref{table:direct-cot-reader-results} compares reader choice (Direct vs CoT Prompting) for Flan-T5-XXL and GPT3. We find that Flan-T5-XXL works better with Direct Prompting as a reader and GPT3 works better with CoT Prompting as a reader. Therefore, for the experiments in the main paper, we go with this choice. Note though that the trends discussed in \S~\ref{sec:exp-results} (\iconsys QA $>$ OneR QA $>$ ZeroR QA) hold regardless of the choice of the reader.

\section{Separate Reader in \iconsys QA}
\label{sec:qa-reader-ablation}

\begin{table}[ht]
    \centering
    \footnotesize
    \setlength{\tabcolsep}{1.0pt}
    \begin{tabular}{p{1.5em}ccccc}\toprule
        & Model &   HotpotQA & 2WikiMQA & MuSiQue & IIRC   \\
        \midrule
        {\multirow{2}{*}{\rotatebox[origin=c]{90}{Flan}}}
        & \fixedicon\sys QA     &  \bf{59.1}\std{0.9}     & \bf{66.5}\std{1.4}     & \bf{30.8}\std{0.2}     & \bf{42.5}\std{2.1} \\
        & w/o reader            &  \nf{52.6}\std{0.3}     & \nf{60.9}\std{0.6}     & \nf{24.9}\std{0.2}     & \nf{40.3}\std{0.2} \\

        \midrule
        {\multirow{2}{*}{\rotatebox[origin=c]{90}{GPT3}}}
        & \fixedicon\sys QA     &  \nf{60.7}\std{1.1}     & \nf{68.0}\std{1.5}     & \bf{36.5}\std{1.2}     & \bf{49.9}\std{1.1} \\
        & w/o reader            &  \bf{61.0}\std{0.7}     & \bf{70.4}\std{1.5}     & \nf{31.5}\std{0.6}     & \nf{48.4}\std{1.0} \\
        \bottomrule
    \end{tabular}
    \caption{Answer F1 of \iconsys QA with and without a separate reader for \texttt{Flan-T5-XXL} (top two rows) and \texttt{GPT3} (bottom two rows). When the reader is not used, the answer is extracted from the CoT generated by \sys while doing the retrieval. Ablating the reader usually hurts the performance.}
    \label{table:qa-reader-ablation}
\end{table}

\iconsys, by construction, produces a CoT as a part of its retrieval process. So, instead of having a separate post-hoc reader, one can also just extract the answer from the CoT generated during retrieval. As Table~\ref{table:qa-reader-ablation} shows the effect of such an ablation.
For \texttt{Flan-T5-XXL} having a separate reader is significantly better. For GPT3, this is not always true, but at least a model with a separate reader is always better or close to the one without. So overall we go with the choice of using the reader for the experiments in this paper.

\section{Prompts}
\label{sec:apndx-prompts}
Our manually written chain-of-thought annotations for HotpotQA, 2WikiMultihopQA, MuSiQue, and IIRC are given in Listing \ref{lst:hotpotqa-cots}, \ref{lst:2wikimultihopqa-cots}, \ref{lst:musique-cots} and \ref{lst:iirc-cots} respectively. Our prompts for GPT3 CoT Prompting are the same as these, except they have Wikipipedia paragraphs on the top of the questions as shown in \S~\ref{subsec:retriever}\footnote{We are not showing the paragraphs in the paper for brevity but they can be obtained from the released code.}. Our prompts for GPT3 Direct Prompting are the same as that of CoT prompting, except have the answer after "A:" directly. Our prompts for Flan-T5-* are slightly different from that of GPT3. For CoT Prompting, we prefix the question line: "Q: Answer the following question by reasoning step-by-step. <actual-question>". For Direct Prompting, we prefix the question line: "Q: Answer the following question. <actual-question>". We did this to follow Flan-T5-*'s training format and found it to help its CoT generation.

\onecolumn
\colorlet{shadecolor}{gray!10}
\UseRawInputEncoding
\lstset{breaklines=true, columns=fullflexible, backgroundcolor=\color{shadecolor}}
\begin{lstlisting}[basicstyle=\footnotesize, title={Chain-of-Thought annotations for HotpotQA.}, caption={Chain-of-Thought annotations for HotpotQA.}, label={lst:hotpotqa-cots}]
Q: Jeremy Theobald and Christopher Nolan share what profession?
A: Jeremy Theobald is an actor and producer. Christopher Nolan is a director, producer, and screenwriter. Therefore, they both share the profession of being a producer. So the answer is: producer.

Q: What film directed by Brian Patrick Butler was inspired by a film directed by F.W. Murnau?
A: Brian Patrick Butler directed the film The Phantom Hour. The Phantom Hour was inspired by the films such as Nosferatu and The Cabinet of Dr. Caligari. Of these Nosferatu was directed by F.W. Murnau. So the answer is: The Phantom Hour.

Q: How many episodes were in the South Korean television series in which Ryu Hye-young played Bo-ra?
A: The South Korean television series in which Ryu Hye-young played Bo-ra is Reply 1988. The number of episodes Reply 1988 has is 20. So the answer is: 20.

Q: Vertical Limit stars which actor who also played astronaut Alan Shepard in "The Right Stuff"?
A: The actor who played astronaut Alan Shepard in "The Right Stuff" is Scott Glenn. The movie Vertical Limit also starred Scott Glenn. So the answer is: Scott Glenn.

Q: What was the 2014 population of the city where Lake Wales Medical Center is located?
A: Lake Wales Medical Center is located in the city of Polk County, Florida. The population of Polk County in 2014 was 15,140. So the answer is: 15,140.

Q: Who was born first? Jan de Bont or Raoul Walsh?
A: Jan de Bont was born on 22 October 1943. Raoul Walsh was born on March 11, 1887. Thus, Raoul Walsh was born the first. So the answer is: Raoul Walsh.

Q: In what country was Lost Gravity manufactured?
A: The Lost Gravity (roller coaster) was manufactured by Mack Rides. Mack Rides is a German company. So the answer is: Germany.

Q: Which of the following had a debut album entitled "We Have an Emergency": Hot Hot Heat or The Operation M.D.?
A: The debut album of the band "Hot Hot Heat" was "Make Up the Breakdown". The debut album of the band "The Operation M.D." was "We Have an Emergency". So the answer is: The Operation M.D..

Q: In which country did this Australian who was detained in Guantanamo Bay detention camp and published "Guantanamo: My Journey" receive para-military training?
A: The Australian who was detained in Guantanamo Bay detention camp and published "Guantanamo: My Journey" is David Hicks. David Hicks received his para-military training in Afghanistan. So the answer is: Afghanistan.

Q: Does The Border Surrender or Unsane have more members?
A: The Border Surrender band has following members: Keith Austin, Simon Shields, Johnny Manning and Mark Austin. That is, it has 4 members. Unsane is a trio of 3 members. Thus, The Border Surrender has more members. So the answer is: The Border Surrender.

Q: Which band formed first, Sponge Cola or Hurricane No. 1?
A: Sponge Cola band was formed in 1998. Hurricane No. 1 was formed in 1996. Thus, Hurricane No. 1 band formed the first. So the answer is: Hurricane No. 1.

Q: James Paris Lee is best known for investing the Lee-Metford rifle and another rifle often referred to by what acronymn?
A: James Paris Lee is best known for investing the Lee-Metford rifle and Lee–Enfield series of rifles. Lee–Enfield is often referred to by the acronym of SMLE. So the answer is: SMLE.

Q: Who was born first, James D Grant, who uses the pen name of Lee Child, or Bernhard Schlink?
A: James D Grant, who uses the pen name of Lee Child, was born in 1954. Bernhard Schlink was born in 1944. Thus, Bernhard Schlink was born first. So the answer is: Bernhard Schlink.

Q: Which American neo-noir science fiction has Pierce Gagnon starred?
A: Pierce Gagnon has starred in One Tree Hill, Looper, Wish I Was Here and Extant. Of these, Looper is an American neo-noir science fiction. So the answer is: Looper.

Q: What year did Edburga of Minster-in-Thanet's father die?
A: The father of Edburga of Minster-in-Thanet is King Centwine. Centwine died after 685. So the answer is: after 685.

Q: Were Lonny and Allure both founded in the 1990s?
A: Lonny (magazine) was founded in 2009. Allure (magazine) was founded in 1991. Thus, of the two, only Allure was founded in 1990s. So the answer is: no.

Q: The actor that stars as Joe Proctor on the series "Power" also played a character on "Entourage" that has what last name?
A: The actor that stars as Joe Proctor on the series "Power" is Jerry Ferrara. Jerry Ferrara also played a character on Entourage named Turtle Assante. Thus, Turtle Assante's last name is Assante. So the answer is: Assante.

Q: Nobody Loves You was written by John Lennon and released on what album that was issued by Apple Records, and was written, recorded, and released during his 18 month separation from Yoko Ono?
A: The album issued by Apple Records, and written, recorded, and released during John Lennon's 18 month separation from Yoko Ono is Walls and Bridges. Nobody Loves You was written by John Lennon on Walls and Bridges album. So the answer is: Walls and Bridges.

Q: How many awards did the "A Girl Like Me" singer win at the American Music Awards of 2012?
A: The singer of "A Girl Like Me" singer is Rihanna. In the American Music Awards of 2012, Rihana won one award. So the answer is: one.

Q: who is older Jeremy Horn or Renato Sobral ?
A: Jeremy Horn was born on August 25, 1975. Renato Sobral was born on September 7, 1975. Thus, Jeremy Horn is older. So the answer is: Jeremy Horn.
\end{lstlisting}
\begin{lstlisting}[basicstyle=\footnotesize, title={Chain-of-Thought annotations for 2WikiMultihopQA.}, caption={Chain-of-Thought annotations for 2WikiMultihopQA.}, label={lst:2wikimultihopqa-cots}]
Q: When did the director of film Hypocrite (Film) die?
A: The film Hypocrite was directed by Miguel Morayta. Miguel Morayta died on 19 June 2013. So the answer is: 19 June 2013.

Q: Do director of film Coolie No. 1 (1995 Film) and director of film The Sensational Trial have the same nationality?
A: Coolie No. 1 (1995 film) was directed by David Dhawan. The Sensational Trial was directed by Karl Freund. David Dhawan's nationality is India. Karl Freund's nationality is Germany. Thus, they do not have the same nationality. So the answer is: no.

Q: Are both Kurram Garhi and Trojkrsti located in the same country?
A: Kurram Garhi is located in the country of Pakistan. Trojkrsti is located in the country of Republic of Macedonia. Thus, they are not in the same country. So the answer is: no.

Q: Who was born first out of Martin Hodge and Ivania Martinich?
A: Martin Hodge was born on 4 February 1959. Ivania Martinich was born on 25 July 1995. Thus, Martin Hodge was born first. So the answer is: Martin Hodge.

Q: Which film came out first, The Night Of Tricks or The Genealogy?
A: The Night of Tricks was published in the year 1939. The Genealogy was published in the year 1979. Thus, The Night of Tricks came out first. So the answer is: The Night Of Tricks.

Q: When did the director of film Laughter In Hell die?
A: The film Laughter In Hell was directed by Edward L. Cahn. Edward L. Cahn died on August 25, 1963. So the answer is: August 25, 1963.

Q: Which film has the director died later, The Gal Who Took the West or Twenty Plus Two?
A: The film Twenty Plus Two was directed by Joseph M. Newman. The Gal Who Took the West was directed by Frederick de Cordova. Joseph M. Newman died on January 23, 2006. Fred de Cordova died on September 15, 2001. Thus, the person to die later from the two is Twenty Plus Two. So the answer is: Twenty Plus Two.

Q: Who is Boraqchin (Wife Of Ögedei)'s father-in-law?
A: Boraqchin is married to Ögedei Khan. Ögedei Khan's father is Genghis Khan. Thus, Boraqchin's father-in-law is Genghis Khan. So the answer is: Genghis Khan.

Q: What is the cause of death of Grand Duke Alexei Alexandrovich Of Russia's mother?
A: The mother of Grand Duke Alexei Alexandrovich of Russia is Maria Alexandrovna. Maria Alexandrovna died from tuberculosis. So the answer is: tuberculosis.

Q: Which film has the director died earlier, When The Mad Aunts Arrive or The Miracle Worker (1962 Film)?
A: When The Mad Aunts Arrive was directed by Franz Josef Gottlieb. The Miracle Worker (1962 film) was directed by Arthur Penn. Franz Josef Gottlieb died on 23 July 2006. Arthur Penn died on September 28, 2010. Thus, of the two, the director to die earlier is Franz Josef Gottlieb, who directed When The Mad Aunts Arrive. So the answer is: When The Mad Aunts Arrive.

Q: Which album was released earlier, What'S Inside or Cassandra'S Dream (Album)?
A: What's Inside was released in the year 1995. Cassandra's Dream (album) was released in the year 2008. Thus, of the two, the album to release earlier is What's Inside. So the answer is: What's Inside.

Q: Are both mountains, Serre Mourene and Monte Galbiga, located in the same country?
A: Serre Mourene is located in Spain. Monte Galbiga is located in Italy. Thus, the two countries are not located in the same country. So the answer is: no.

Q: What is the date of birth of the director of film Best Friends (1982 Film)?
A: The film Best Friends was directed by Norman Jewison. Norman Jewison was born on July 21, 1926. So the answer is: July 21, 1926.

Q: Which film has the director born first, Two Weeks With Pay or Chhailla Babu?
A: Two Weeks with Pay was directed by Maurice Campbell. Chhailla Babu was directed by Joy Mukherjee. Maurice Campbell was born on November 28, 1919. Joy Mukherjee was born on 24 February 1939. Thus, from the two directors, Chhailla Babu was born first, who directed Two Weeks With Pay. So the answer is: Two Weeks With Pay.

Q: Who is the grandchild of Krishna Shah (Nepalese Royal)?
A: Krishna Shah has a child named Rudra Shah. Rudra Shah has a child named Prithvipati Shah. Thus, Krishna Shah has a grandchild named Prithvipati Shah. So the answer is: Prithvipati Shah.

Q: When was the director of film P.S. Jerusalem born?
A: P.S. Jerusalem was directed by Danae Elon. Danae Elon was born on December 23, 1970. So the answer is: December 23, 1970.

Q: Which album was released more recently, If I Have to Stand Alone or Answering Machine Music?
A: If I Have to Stand Alone was published in the year 1991. Answering Machine Music was released in the year 1999. Thus, of the two, the album to release more recently is Answering Machine Music. So the answer is: Answering Machine Music.

Q: Where did the director of film Maddalena (1954 Film) die?
A: The film Maddalena is directed by Augusto Genina. Augusto Genina died in Rome. So the answer is: Rome.

Q: When did the director of film The Boy And The Fog die?
A: The director of The Boy and the Fog is Roberto Gavaldón. Roberto Gavaldón died on September 4, 1986. So the answer is: September 4, 1986.

Q: Are the directors of films The Sun of the Sleepless and Nevada (1927 film) both from the same country?
A: The director of Sun of the Sleepless is Temur Babluani. The director of Nevada (1927 film) is John Waters. John Waters is from the country of America. Temur Babluani is from the country of Georgia. Thus, John Walters and Temur Babluani are not from the same country. So the answer is: no.
\end{lstlisting}
\begin{lstlisting}[basicstyle=\footnotesize, title={Chain-of-Thought annotations for MuSiQue.}, caption={Chain-of-Thought annotations for MuSiQue.}, label={lst:musique-cots}]
Q: When did the first large winter carnival take place in the city where CIMI-FM is licensed to broadcast?
A: CIMI-FM is licensed to broadcast in Quebec City. The first large winter carnival in Quebec City took place in 1894. So the answer is: 1894.

Q: When was Neville A. Stanton's employer founded?
A: The employer of Neville A. Stanton is University of Southampton. The University of Southampton was founded in 1862. So the answer is: 1862.

Q: What county is Hebron located in, in the same province the Heritage Places Protection Act applies to?
A: Heritage Places Protection Act applies to the jurisdiction of Prince Edward Island. Hebron, Prince Edward Island is located in the Prince County. So the answer is: Prince County.

Q: What weekly publication in the Connecticut city with the most Zagat rated restaurants is issued by university of America-Lite: How Imperial Academia Dismantled Our Culture's author?
A: The author of America-Lite: How Imperial Academia Dismantled Our Culture is David Gelernter. David Gelernter was educated at the Yale University. The city in Connecticut that has the highest number of Zagat-rated restaurants is New Haven. The weekly publication in New Haven that is issued by Yale University is Yale Herald. So the answer is: Yale Herald.

Q: What is the headquarters for the organization who sets the standards for ISO 21500?
A: The standards for ISO 21500 were set by International Organization for Standardization. The International Organization for Standardization has headquarters in Geneva. So the answer is: Geneva.

Q: What did the publisher of Banjo-Tooie rely primarily on for its support?
A: The publisher of Banjo-Tooie is Nintendo. Nintendo relied primarily for its support on first-party games. So the answer is: first-party games.

Q: In which county was the birthplace of the Smoke in tha City performer?
A: The performer of Smoke in tha City is MC Eiht. MC Eiht's birthplace is Compton. Compton is located in the county of Los Angeles County. So the answer is: Los Angeles County.

Q: What region of the state where Guy Shepherdson was born, contains SMA Negeri 68?
A: Guy Shepherdson was born in Jakarta. SMA Negeri 68 Jakarta is located in Central Jakarta. So the answer is: Central Jakarta.

Q: When did Britain withdraw from the country containing Hoora?
A: Hoora is in the country of Bahrain. Britain withdrew from Bahrain in 1971. So the answer is: 1971.

Q: Where does the Snake River start, in the state where Lima Mountain is located?
A: Lima Mountain is located in the state of Minnesota. The snake river in Minnesota starts in southern Aitkin County. So the answer is: southern Aitkin County.

Q: What shares a border with Rivière-Verte in the province WRSU-FM broadcasts in?
A: WRSU-FM was licensed to broadcast to New Brunswick. Rivière-Verte, New Brunswick shares border with Edmundston. So the answer is: Edmundston.

Q: When was the state of emergency declared in the country where the Senate is located?
A: The Senate is in the country of Kenya. The state of emergency was declared in Kenya on 20 October 1952. So the answer is: 20 October 1952.

Q: How long is the US border with the country that borders the state where Finding Dory takes place?
A: Finding Dory is supposed to take place in California. The country that shares a border with California is Mexico. The length of the us border with Mexico is 1,989 mi. So the answer is: 1,989 mi.

Q: What genre is the record label of the performer of So Long, See You Tomorrow associated with?
A: The performer of So Long, See You Tomorrow is Bombay Bicycle Club. The record label of Bombay Bicycle Club is Island Records. The genre of Island Records is jazz. So the answer is: jazz.

Q: When did the first large winter carnival happen in Olivier Robitaille's place of birth?
A: Olivier Robitaille was born in Quebec City. The first large winter carnival in Quebec City happened in the 1894. So the answer is: 1894.

Q: What is the genre of the record label of the band that performed on the Crush Tour?
A: The Crush Tour is performed by the band Bon Jovi. The record label of Bon Jovi is Island Records. The genre of Island Records is jazz. So the answer is: jazz.

Q: When was the first railway line constructed between Kotri and the city where Marie Adelaide Leprosy Centre is located?
A: Marie Adelaide Leprosy Centre is located in Karachi. The first railway line between Kotri and Karachi was constructed in April 1858. So the answer is: April 1858.

Q: Where is the crying stone found in the country in which Raphael Tuju holds citizenship?
A: Raphael Tuju is a citizen of Kenya. The crying stone in Kenya is found along the highway towards Kisumu. So the answer is: along the highway towards Kisumu.

Q: When did Britain withdraw from the country where the village of Wadyan is found?
A: Wadyan is in the country of Bahrain. Britain withdraw from Bahrain in 1971. So the answer is: 1971.

Q: How many countries in Pacific National University's continent are recognized by the organization that mediated the truce ending the Iran-Iraq war?
A: Pacific National University is located in Khabarovsk, Russia Khabarovsk, Russian is in the continent of Asia. The entity that mediated the truce which ended the Iran-Iraq War is the UN. The number of member states that UN recognises in Asia is 53. So the answer is: 53.
\end{lstlisting}
\begin{lstlisting}[basicstyle=\footnotesize, title={Chain-of-Thought annotations for IIRC.}, caption={Chain-of-Thought annotations for IIRC.}, label={lst:iirc-cots}]
Q: What is the age difference between the kicker and the quarterback for the Chargers?
A: The kicker for the Chargers is Nate Kaeding. The quarterback (QB) for the Chargers is Philip Rivers. Nate Kaeding was born in the year 1982. Philip Rivers was born in the year 1981. Thus, the age difference between them is of 1 year. So the answer is: 1.

Q: How many years was the ship that took the battalion from New South Wales to Ceylon in service?
A: The ship that took the battalion from New South Wales to Ceylon is General Hewitt. General Hewitt was launched in Calcutta in 1811. General Hewitt was sold for a hulk or to be broken up in 1864. So she served for a total of 1864 - 1811 = 53 years. So the answer is: 53.

Q: What year was the theatre that held the 2016 NFL Draft built?
A: The theatre that held the 2016 NFL Draft is Auditorium Theatre. The Auditorium Theatre was built in 1889. So the answer is: 1889.

Q: How long had Milan been established by the year that Nava returned there as a reserve in the first team's defense?
A: Nava returned to Milan as a reserve in the first team's defense in the year 1990. Milan had been established in the year 1899. Thus, Milan had been established for 1990 - 1899 = 91 years when Milan returned to Milan as a reserve in the first team's defense. So the answer is: 91.

Q: When was the town Scott was born in founded?
A: Scott was born in the town of Cooksville, Illinois. Cooksville was founded in the year 1882. So the answer is: 1882.

Q: In what country did Wright leave the French privateers?
A: Wright left the French privateers in Bluefield's river. Bluefields is the capital of the South Caribbean Autonomous Region (RAAS) in the country of Nicaragua. So the answer is: Nicaragua.

Q: Who plays the A-Team character that Dr. Hibbert fashioned his hair after?
A: Dr. Hibbert fashioned his hair after Mr. T from The A-Team. Mr T.'s birthname is Lawrence Tureaud. So the answer is: Lawrence Tureaud.

Q: How many people attended the conference held near Berlin in January 1942?
A: The conference held near Berlin in January 1942 is Wannsee Conference. Wannsee Conference was attended by 15 people. So the answer is: 15.

Q: When did the country Ottwalt went into exile in founded?
A: Ottwalt went into exile in the country of Denmark. Denmark has been inhabited since around 12,500 BC. So the answer is: 12,500 BC.

Q: When was the J2 club Uki played for in 2001 founded?
A: The J2 club that Uki played for is Montedio Yamagata. Montedio Yamagata was founded in 1984. So the answer is: 1984.

Q: When was the person who produced A Little Ain't Enough born?
A: A Little Ain't Enough was produced by Bob Rock. Bob Rock was born on April 19, 1954. So the answer is: April 19, 1954.

Q: Which of the schools Fiser is affiliated with was founded first?
A: The schools that Fiser is affiliated with (1) Academy of Music, University of Zagreb (2) Mozarteum University of Salzburg (3) Croatian Music Institute orchestra. Academy of Music, University of Zagreb was founded in the year 1829. Mozarteum University of Salzburg was founded in the year 1841. Croatian Music Institute was founded in the year 1827. Thus, the school founded earliest of these is Croatian Music Institute. So the answer is: Croatian Music Institute.

Q: How many casualties were there at the battle that Dearing fought at under Jubal Early?
A: Under Jubal Early, Dearing fought the First Battle of Bull Run. First Battle of Bull Run has 460 union casualties and 387 confederate casualties. Thus, in total the First Battle of Bull Run had 460 + 387 = 847 casualties. So the answer is: 847.

Q: Which of the two congregations which provided leadership to the Pilgrims was founded first?
A: The congregations which provided leadership to the Pilgrims are Brownists and Separatist Puritans. Brownist was founded in 1581. The Separatist Puritans was founded in 1640. Thus, Brownist was founded first. So the answer is: Brownist.

Q: How long had the Rock and Roll Hall of Fame been open when the band was inducted into it?
A: The band was inducted into Rock and Roll Hall of Fame in the year 2017. Rock and Roll Hall of Fame was established in the year of 1983. Thus, Rock and Roll Hall of Fame been open for 2018 - 1983 = 34 years when the band was inducted into it. So the answer is: 34.

Q: Did the Lord Sewer who was appointed at the 1509 coronation live longer than his king?
A: Lord Sewer who was appointed at the 1509 coronation was Robert Radcliffe, 1st Earl of Sussex. Lord Sever's king in 1509 was Henry VIII of England. Robert Radcliffe, 1st Earl of Sussex was born in the year 1483, and died in the year 1542. So Robert lived for 1542 - 1483 = 59 years. Henry VIII of England was born in the year 1491 and died in the year 1547. So Henry VIII lived for 1547 - 1491 = 56 years. Thus, Robert Radcliffe lived longer than Henry VIII. So the answer is: yes.

Q: When was the place near where Manuchar was defeated by Qvarqvare established?
A: Manuchar was defeated by Qvarqvare near Erzurum. Erzurum was founded during the Urartian period. So the answer is: Urartian period.

Q: What year was the man who implemented the 46 calendar reform born?
A: The man who implemented the 46 calendar reform is Julius Caesar. Julius Caesar was born in the year 100 BC. So the answer is: 100 BC.

Q: How many years after the first recorded Tommy John surgery did Scott Baker undergo his?
A: The first recorded Tommy John surgery happened when it was invented in the year 1974. Scott Baker underwent Tommy John surgery in the year 2012. Thus, Scott Baker underwent Tommy John surgery 2012 - 1974 = 38 years after it was first recorded. So the answer is: 38.

Q: Which was the older of the two players who found the net in the Double-Headed Eagle of the North in the sixth final for PAOK?
A: The two players who found the net in the Double-Headed Eagle of the North in the sixth final for PAOK are Koudas and Matzourakis. Koudas was born on 23 November 1946. Matzourakis was born on 6 June 1949. Thus, the older person among the two is Koudas. So the answer is: Koudas.
\end{lstlisting}
\twocolumn

\end{document}